\documentclass{article}

\usepackage{PRIMEarxiv}

\usepackage[utf8]{inputenc} 
\usepackage[T1]{fontenc}    
\usepackage{hyperref}       
\usepackage{url}            
\usepackage{booktabs}       
\usepackage{amsfonts}       
\usepackage{nicefrac}       
\usepackage{microtype}      
\usepackage{lipsum}
\usepackage{multirow}  
\usepackage{amsmath,amssymb,amsfonts}
\usepackage{subcaption}
\usepackage{xcolor}
\usepackage{fancyhdr}       
\usepackage{graphicx}       
\usepackage{hyperref}
\graphicspath{{media/}}     
\usepackage{perpage}
\MakePerPage{footnote}
\usepackage{appendix}
\usepackage{minted}

\makeatletter 
\def\thanks#1{\protected@xdef\@thanks{\@thanks \protect\footnotetext{#1}}} \makeatother 

\title{Advancing Automated Algorithm Design via Evolutionary Stagewise Design with LLMs
}

\author{
  Chen Lu$^{1,2}$, Ke Xue$^{1,2}$, Chengrui Gao$^{1,2}$, Yunqi Shi$^{1,2}$, Siyuan Xu$^{3}$, Mingxuan Yuan$^{3}$, \\
  \textbf{Chao Qian}$^{\dagger1,2}$~\textbf{,} \textbf{Zhi-Hua Zhou}$^{\dagger1,2}$
\thanks{$\dagger$ Corresponding Author: \texttt{\{qianc,zhouzh\}@nju.edu.cn}} \\
  $^1$ National Key Laboratory for Novel Software Technology, Nanjing University, China\\
  $^2$ School of Artificial
Intelligence, Nanjing University, China\\
  $^3$ Huawei Noah’s Ark Lab, China\\
}

\begin{document}
\maketitle

\begin{abstract}
With the rapid advancement of human science and technology, problems in industrial scenarios are becoming increasingly challenging, bringing significant challenges to traditional algorithm design, which is typically tedious and inefficient, with a high dependence on expertise. Automated algorithm design with large language models (LLMs) emerges as a promising solution, but the currently adopted black-box modeling deprives LLMs of any awareness of the intrinsic mechanism of the target problem, leading to hallucinated designs under industrial scenarios with limited evaluation budgets. In this paper, we introduce Evolutionary Stagewise Algorithm Design (EvoStage), a novel evolutionary paradigm that bridges the gap between the rigorous demands of industrial-scale algorithm design and the LLM-based algorithm design methods. Drawing inspiration from chain-of-thought, EvoStage decomposes the algorithm design process into sequential, manageable stages and integrates real-time intermediate feedback to iteratively refine algorithm design directions. To further reduce the algorithm design space and avoid falling into local optima, we introduce a multi-agent system and a "global-local perspective" mechanism, enabling the system to coordinate different LLM agents to design different algorithm components and balance the local stage optimization and the global overall optimization. We apply EvoStage to the design of two types of common optimizers (i.e., gradient-based and black-box optimizers): designing parameter configuration schedules of the Adam optimizer for chip placement, and designing acquisition functions of Bayesian optimization for black-box optimization. Experimental results across open-source benchmarks demonstrate that EvoStage outperforms human-expert designs and existing LLM-based methods within only a couple of evolution steps, even achieving the historically state-of-the-art half-perimeter wire-length results on every tested chip case. Furthermore, when deployed on a commercial-grade 3D chip placement tool, EvoStage significantly surpasses the original performance metrics (e.g., with a 9.24\% improvement of the half-perimeter wirelength on the logic die), achieving record-breaking efficiency with a 52.21\% improvement of the optimization iterations. We hope EvoStage can significantly advance automated algorithm design in the real world, helping elevate human productivity.

\end{abstract}


\section{Introduction}

Algorithm design lies at the heart of problem-solving. In conventional approaches of problem-solving, experts iteratively improve algorithms by synthesizing domain knowledge with feedback from previous iterations, following a hierarchical and progressive workflow: a) Deploy off-the-shelf algorithms; b) Adapt existing algorithms to new cases via adjusting their configurations; c) Make novel designs to specific components of the existing algorithms, tailoring them to the target cases. However, this process is excessively tedious and inefficient, with a high dependence on expertise, severely restricting the efficiency of solving real-world complex problems and thus hindering human productivity. For example, the notably high threshold for chip placement arises from the fact that the task requires prolonged iterative improvement across the entire lengthy upstream and downstream pipelines by experts with extensive professional experience. Such reliance on expertise underscores the urgent need for a new, more efficient, and automated paradigm in industrial algorithm design.

Recently, large language models (LLMs) have demonstrated expert-level capabilities in areas like logical reasoning~\cite{deepseek-r1} and code generation~\cite{deepseek-coder}, thus holding great potential for designing algorithms automatically. In particular, there has been an increasing emphasis on automated algorithm design leveraging LLMs in an evolutionary framework~\cite{alphaevolve, EoH, FunSearch, ReEvo}. The evolutionary framework is derived from classic evolutionary algorithms inspired by Darwin's theory of evolution~\cite{back1996evolutionary,elbook}, where we maintain a population of algorithms (or heuristics which refer to key components of an algorithm) produced by the LLMs, with critical information about the algorithms (e.g., the performance scores or the natural language descriptions about them~\cite{EoH}), and select parent algorithms in certain strategies as references for the LLMs to produce next-generation algorithms. This paradigm was first proposed by Google as FunSearch~\cite{FunSearch}, based on which a system called AlphaEvolve~\cite{alphaevolve} has been recently constructed, where a large population of diverse algorithms is maintained in the form of computer programs with an island-based evolutionary framework, encouraging exploration and avoiding local optima. EoH~\cite{EoH} evolves the heuristics (represented by codes) along with the natural language descriptions about them, providing the LLMs with more understanding of the heuristics. ReEvo~\cite{ReEvo} additionally utilizes an LLM reflector to give verbal reflections for better generation of heuristics. These methods have been applied to address classical combinatorial optimization problems and make new discoveries for open mathematical problems, yielding algorithms with excellent performance. Such a paradigm of automated algorithm design by evolving algorithms with LLMs has even been specified as a promising way for realizing the Level 4 agentic systems (i.e., the self-evolving agents) in Google's white paper~\cite{whitepaper}.

With the rapid advancement of human science and technology, real-world problems, e.g., optimization problems in industrial scenarios such as chip placement~\cite{chu2009placement}, have become increasingly intractable and complex. They typically exhibit the following two characteristics: \textbf{a) Expensive evaluation~\cite{expensivebbo}.} It often takes a long time or a huge cost to evaluate the quality of a solution or the performance of an algorithm, which implies a limited evaluation budget for algorithm design. For example, in chip placement, it usually takes several hours to evaluate the power, performance, and area (PPA) of a concrete placement result~\cite{graph-placement}. \textbf{b) Insufficient high-quality samples with significant gaps between case instances.} We often have no access to large-scale datasets consisting of high-quality samples, and moreover, there are often substantial discrepancies between case instances, making it difficult to devise universally applicable solutions. For example, for industrial-grade chips, design details are often kept confidential as proprietary business information, and due to the differences in functionality as well as in type and amount of integrated devices, chip cases typically exhibit substantial discrepancies. These characteristics pose significant challenges to automated algorithm design, which require us to adaptively generate a tailored algorithm for each individual problem instance using a limited budget of real evaluations, while without successful experiences from prior cases.

Although automated algorithm design with LLMs has emerged as a revolutionary paradigm shift in the field of algorithm research and development, holding great potential to subversively elevate human productivity, the current methods introduced above cannot meet the demand of algorithm design in industry. All of them have adopted a simple black-box modeling for algorithm design tasks~\cite{alphaevolve, EoH, FunSearch, ReEvo}, where LLMs only receive simple reward feedback after the overall completion of algorithm design to guide the evolution of algorithm design in subsequent rounds. Such black-box modeling deprives LLMs of any awareness of the intrinsic mechanism of the target problem, disabling them from utilizing intermediate feedback to correct possibly wrong design directions and consequently costing these methods numerous evaluations to ultimately derive a high-quality algorithm design~\cite{FunSearch}, which is often infeasible in real-world industrial scenarios with limited evaluation budgets. Moreover, the existing approaches have primarily focused on algorithm design for well-studied and well-established problems (e.g., common combinatorial optimization problems~\cite{EoH, ReEvo, FunSearch}, search for open mathematical problems~\cite{alphaevolve}, etc). These problems possess high-quality sample data in the pre-training corpus of LLMs, enabling LLMs to have a comprehensive grasp of these problems and consequently simplifying the corresponding algorithm design process. In contrast, problems in real-world industrial scenarios typically lack high-quality sample data and exhibit significant case-by-case variations. Therefore, LLMs may not have encountered similar settings during pre-training, which further exacerbates the hallucination~\cite{sourcesofhallucination} of LLMs: LLMs tend to apply existing mismatched knowledge~\cite{llmlies} to optimize algorithm design, which appears logically coherent but may lead to completely incorrect optimization directions in practice.

\begin{figure}[t]
\centering
    \centering
    \includegraphics[width=\textwidth]{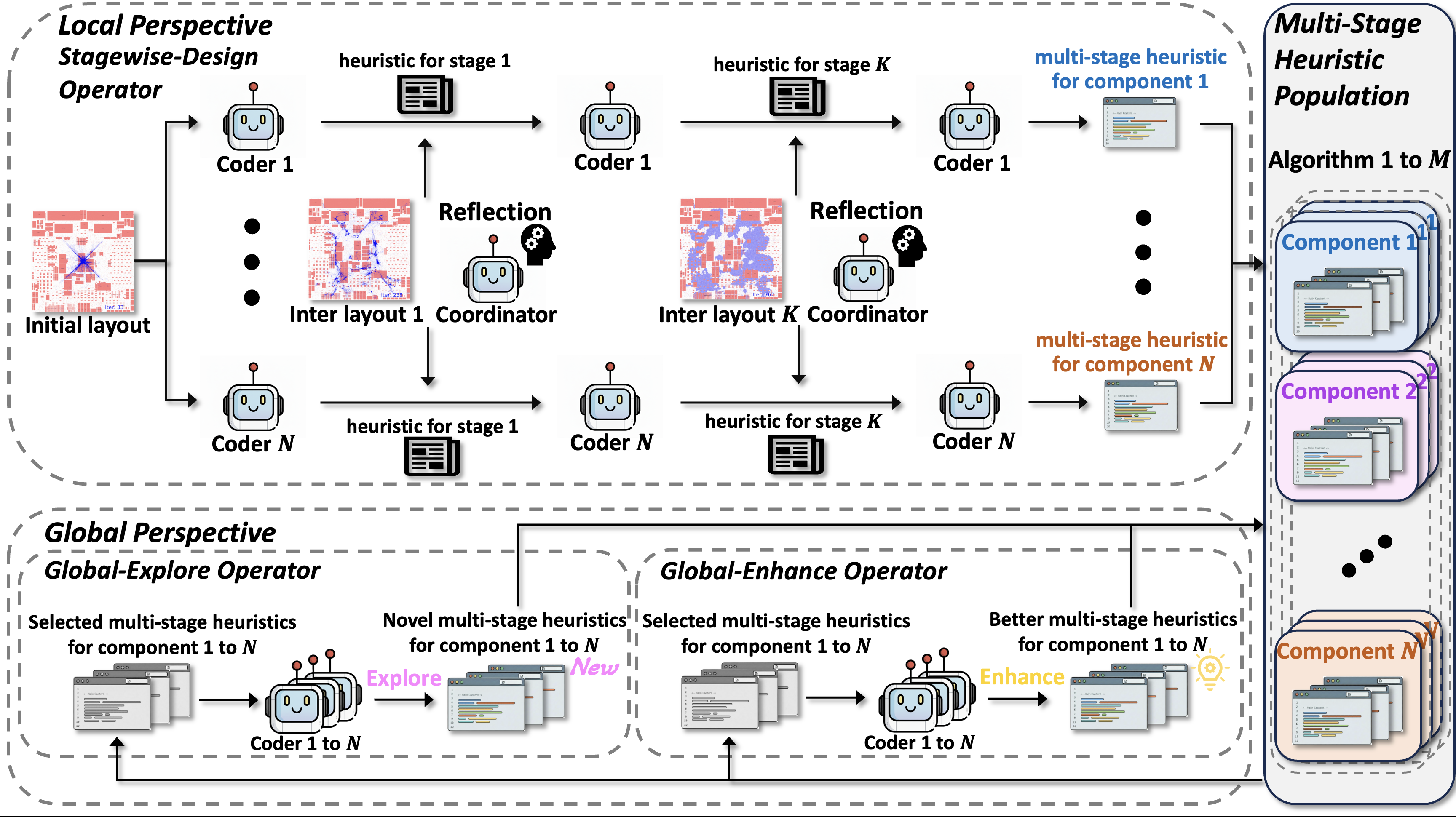} 
\caption{The overview of EvoStage. The upper left part is the intuitive illustration of the local perspective operator (i.e., the Stagewise-Design operator), where the algorithm design task is decomposed automatically by the LLM coordinator agent, with the coordinator reflecting on the intermediate information (denoted as "Inter layout", using chip placement as an example) and giving guidance for the next stage, and $N$ LLM coder agents producing $N$ algorithm components (one coder responsible for one respective component) according to the guidance from the coordinator stage-by-stage. The lower left part is the intuitive illustration of the two global perspective operators. The Global-Explore operator is to prompt the coder to produce a novel multi-stage heuristic in one shot with a selected reference, while the Global-Enhance operator is to prompt the coder to produce a better multi-stage heuristic in one shot with a selected reference. The right part is the multi-stage heuristic population, which maintains $M$ algorithms. Each algorithm is divided into $N$ components, each of which is implemented as a multi-stage heuristic. When selecting an algorithm from the population, we select the corresponding $N$ multi-stage heuristics for the $N$ components and assemble a complete algorithm from them. EvoStage performs an iterative process of selecting parent algorithms from the population, reproducing new offspring algorithms by operators, and updating the population to keep better algorithms, with the goal of iteratively improving the algorithms maintained in the population and finally generating a high-performing algorithm.}
\label{fig:overview}
\end{figure}

In this work, we seek to advance automated algorithm design with LLMs, enabling it even for unseen industry-level scenarios with a limited evaluation budget. Drawing inspiration from the remarkable reasoning enhancement of LLMs via chain-of-thought (CoT)~\cite{CoT-origin}, and building on the empirical evidence that supplementary domain knowledge can boost response accuracy and alleviate hallucinations in LLMs~\cite{RAG-origin}, we propose Evolutionary Stagewise Algorithm Design (EvoStage), a new paradigm that better aligns with the capabilities of LLMs and boosts LLMs' performance in the industry-level algorithm design tasks with limited budgets. Specifically, EvoStage decomposes the challenging algorithm design task into multiple stages, with each stage as a simpler subtask. By solving the subtasks sequentially, the LLM agents finally accomplish the overall algorithm design stage-by-stage, generating multi-stage heuristics. In each subtask, the LLM agents are required to carry out algorithm design for the corresponding stage, and they can receive real-time feedback on algorithm-execution outcomes of the current stage. This intermediate feedback updates LLM agents' domain-specific understanding of the target problem, which can help mitigate the hallucinations of LLM-designed algorithms and improve their performance and reliability. Through the evolution of stagewise designs, the LLM agents develop a more comprehensive and in-depth understanding of the domain knowledge and characteristics of the target problem, ultimately leading to the proposal of novel and high-quality algorithms. Moreover, to reduce the complexity of the algorithm design space and ensure a consensus among individual LLM coder agents, we construct a multi-agent system to tap into the full potential of LLMs in the proposed paradigm, with one LLM coder agent for one algorithm component (also called heuristic) to be designed, as well as one LLM coordinator agent for reflecting on stage information and offering the goal and guidance for the next stage to the LLM coder agents. In order to prevent LLM agents from overfocusing on subtask optimization and falling into local optima, guided by recent theoretical advances in evolutionary learning~\cite{elbook}, we also propose a "global-local perspective" mechanism analogous to "fast and slow thinking"~\cite{chen2025pangu}, which balances the global optimization of the overall algorithm and the local optimization of subtasks through the global perspective operator (i.e., prompting the LLMs to analyze the multi-stage heuristic references and generate new ones in one shot) and local perspective operator (i.e., prompting the LLMs to design algorithm components stage-by-stage). An overview of EvoStage is shown in Figure~\ref{fig:overview}.

We demonstrate the outstanding performance of the proposed EvoStage for optimizer design in industry-level scenarios. We first apply EvoStage to design the learning rate schedules as well as the optimization step schedules for the popular gradient-based optimizer Adam~\cite{kingma2014adam} in the optimization of global placement (GP)~\cite{industrialeda}. GP is a fundamental step in the physical design flow of VLSI circuits~\cite{chu2009placement}, which provides a rough overall layout of the cells, and thus has a direct and significant impact on the PPA of the final chip design~\cite{lin2020dreamplace}. EvoStage surpasses the historically state-of-the-art GP results obtained by different placers (i.e., DREAMPlace~\cite{lin2020dreamplace} and Xplace~\cite{xplace}) that are elaborately designed and fine-tuned by human experts on all the chip cases of two open-source benchmarks~\cite{ispd2005,iccad2015} within only 25 evaluations, as well as beating state-of-the-art LLM-based algorithm design methods (i.e., AlphaEvolve~\cite{alphaevolve} and EoH~\cite{EoH}). What is even more inspiring, we have applied EvoStage to a commercial-grade 3D chip placement tool, significantly outperforming the original results across multiple metrics and achieving dramatic efficiency on a real-world industrial 3D chip design with a 9.24\% improvement of the half-perimeter wirelength on the logic die and a dramatic 52.21\% improvement of the optimization iterations. 

It is well known that optimization is generally categorized into two classes: gradient-based optimization and black-box optimization (which refers to the scenario where gradient information is inaccessible, and has wide applications in science and engineering, such as drug discovery~\cite{terayama2021black} and material design~\cite{frazier2015bayesian}). Adam~\cite{kingma2014adam}, as mentioned earlier, is a representative gradient-based optimizer, while for black-box optimization, Bayesian optimization (BO)~\cite{BOsurvey} represents one of the most widely used approaches, which leverages a probabilistic model (typically a Gaussian process) to capture the uncertainty of the unknown objective function, then uses an acquisition function to select the next most "promising" point with the guidance of the model. In order to verify the broad generalizability of our proposed EvoStage, we extend its application to the acquisition function design for the widely-used black-box optimization algorithm BO. EvoStage achieves excellent performance on open-source benchmarks, including popular synthetic functions with diverse characteristics that simulate the complex real-world problems, as well as neural architecture search problems~\cite{nas}, surpassing traditional expert-designed acquisition functions and those designed by other LLM-based algorithm design methods.





Our proposed EvoStage has advanced automated algorithm design via stage-wise algorithm design and evolving multi-stage heuristics with LLMs. Empirical results demonstrate that EvoStage exhibits exceptional efficiency and algorithm design capabilities that surpass those of human experts, even for complex problems like chip placement in industrial scenarios. EvoStage holds great potential to facilitate fully automated algorithm design with superior performance over human-designed ones in a highly efficient data-driven manner, accomplishing tasks that are difficult or even impossible for human experts to accomplish. Hopefully, with further development, the large-scale application of automated algorithm design in industry will substantially elevate human productivity. Moreover, our research opens up a new valuable application scenario (beyond conventional applications such as chatbots and coding assistants) for LLMs, which possesses great practical utility with broad demand in the real world.



\section{Method}
\label{sec:method}

\paragraph{Algorithm overview} EvoStage consists of one basic evolutionary framework and three novel components: 
\begin{itemize}
    \item Evolutionary framework, which maintains a population of algorithms and uses a selection strategy to provide the LLMs with references for reproducing new better algorithms by some operators. This basic framework can be viewed as an effective test-time scaling method~\cite{tts}, making LLMs able to focus on the important information and more critical parts in the long context of the algorithm design task. For example, when prompting the LLM to generate algorithms with better performance, the evolutionary framework can provide the LLM with the current best algorithm as a reference and place less emphasis on earlier failed attempts.
    \item Stagewise Design, which is the core paradigm to decompose the algorithm design task into multiple stages and assist the LLMs to solve the task stage-by-stage, better aligning with the capabilities of LLMs and boosting LLMs’ performance in algorithm design.
    \item Multi-agent system, which assigns each LLM coder agent the design task of an algorithm component and arranges a coordinator to perform reflection on the current stage information, cutting down the design space of the algorithm, and coordinating different agents to keep the consistency of the overall design.
    \item Global-local perspective mechanism, which contains global perspective operators to analyze the multi-stage heuristic references and generate new multi-stage heuristics in one shot, helping the LLMs escape local optima in algorithm design, as well as a local perspective operator to generate multi-stage heuristics stage-by-stage.
\end{itemize}

Formally, we consider algorithms computable by Turing machines~\cite{turingmachine}, denoted as the algorithm space $\mathcal{A}$. As Turing machines execute algorithms through state transitions, each algorithm $\boldsymbol{A}\in \mathcal{A}$ can be divided into $K$ stages $\boldsymbol{A}=(A_1, A_2, \dots A_K)$, with each stage containing several Turing machine states. As many algorithms have multiple components, we define an algorithm with $N$ components as $\boldsymbol{A}=(\boldsymbol{C}_1, \boldsymbol{C}_2, \dots, \boldsymbol{C}_N)$, with each component has $K$ stages $\boldsymbol{C}_i=(C_{i,1}, C_{i,2}, \dots, C_{i,K})$, and therefore, we have the partial algorithm at stage $i$ as $A_i=(C_{1,i}, C_{2,i}, \dots, C_{N,i})$. We further define the information about the overall execution of algorithm $\boldsymbol{A}$ as $\boldsymbol{I}=(I_1, I_2, \dots,I_K)$ with $K$ stages, which contains all critical information about the corresponding algorithm (e.g., intermediate information during execution, final performance, etc). The goal of the algorithm design task is to find the best algorithm $\boldsymbol{A}^*$ from the algorithm space $\mathcal{A}$ that maximizes a score function:$\boldsymbol{A}^* = \mathop{\arg \max}_{\boldsymbol{A}\in \mathcal{A}} \;\text{score}(\boldsymbol{A}),$ where the score function is used to measure the quality of the designed algorithm.

\subsection{Evolutionary framework}\label{sec-framework}

The evolutionary framework, inspired from Darwin's theory of evolution, simulates the evolutionary mechanism of natural organisms (i.e., natural selection and genetic variation), which has been widely used for optimization and learning~\cite{back1996evolutionary,elbook}. Here, we employ it for algorithm design, which maintains a population of algorithms and iteratively improves them by selection and reproduction. In the evolutionary framework, there are three critical components: population, selection strategy, and reproduction operators. 

\paragraph{Population} As shown in the right part of Figure~\ref{fig:overview}, in the population, we maintain a pool of $M$ algorithms produced by the LLMs, with additional information about the algorithms (e.g., the performance scores of the algorithms and the natural language descriptions about the them): $\mathcal{A}_{\text{pop}}:=\{(\boldsymbol{A}_i,\boldsymbol{I}_i)\}_{i=1}^M$, where $\boldsymbol{A}_i$ and $\boldsymbol{I}_i$ denote the $i$-th algorithm in the population and its corresponding information, respectively, and $M$ is the population size. Note that we divide a complete algorithm into $N$ components, and each algorithm component is implemented as a multi-stage heuristic, which will be introduced in detail later. Thus, the population is also named as the multi-stage heuristic population, which can be seen as a dynamic database produced by the LLMs during the iterative algorithm generation process. 

\paragraph{Selection} The selection strategy is used to select promising algorithms in the current population as appropriate references for the LLMs to generate new algorithms: $\boldsymbol{A}_{\text{ref}}=\text{select}(\mathcal{A}_{\text{pop}})$. One simple example is always to provide the algorithm with the currently highest score, but this strategy may lead to local optima, as it ignores the diversity of the algorithms, and currently bad-performing algorithms may also have great potential if modified appropriately. In practice, an arbitrary selection strategy can be applied, and we recommend a strategy~\cite{EoH} that selects an algorithm $\boldsymbol{A}$ with probability $p \propto 1 / (r + M)$, where $r$ is its performance rank and $M$ is the population size. In this strategy, algorithms with better performance are more likely to be selected, and underperforming algorithms still have some chance to be selected, which balances the exploration and exploitation. 

\paragraph{Operator} The evolutionary operators are used to prompt the LLMs to produce promising algorithms with the selected references: $\boldsymbol{A}_\text{new}=\text{LLMs}(\text{Op},\{\boldsymbol{A}^i_\text{ref}, \boldsymbol{I}^i_\text{ref}\}^k_{i=1})$ where $\text{Op}$ denotes the operator prompt and $k$ is the number of selected references. For example, in the early stage, we can use operators that encourage the LLMs to produce more diverse algorithms different from the current references, while in the final stage, we can use operators that ask the LLMs to enhance the currently best-performing algorithms by slightly modifying them. The design of the operators will have a significant impact on the overall algorithm generation quality, and their implementation in our framework will be introduced in detail in Section~\ref{operator}. 

After generating new algorithms by selection and operator, the population will be updated according to a certain criterion (e.g., score ranking or diversity) to provide more promising references for the LLMs in the later algorithm generation process. Thus, the evolutionary framework can push the boundaries of the LLM prompting to a new level by providing the LLMs with delicately selected references and guidance on the improving strategy to shift the focus of LLMs to the more important information in the long context of the algorithm design task to produce more promising algorithms. In our implementation, we simply keep the $M$ algorithms with top-$M$ scores in the population update.

\subsection{Stagewise Design}\label{sec-MSH}
Inspired by the surprising improvement in reasoning ability elicited by CoT~\cite{CoT-origin}, we propose Stagewise Design, a new paradigm that better aligns with the capabilities of LLMs to perform multi-step reasoning~\cite{multi-step} and boosts the performance of LLMs in algorithm design. The core insight behind CoT~\cite{CoT-origin} is that when confronted with complex reasoning problems that rarely have analogous cases in pre-training datasets, LLMs cannot directly yield solutions, but instead, they can decompose the target reasoning problem into intermediate reasoning steps, perform stepwise inference on each, and ultimately derive the correct reasoning outcome for the target problem. Therefore, we similarly decompose the complex algorithm design task into multiple intermediate stages, with each stage as a simpler design subtask. In each subtask, the LLM agents are required to carry out algorithm design for the corresponding algorithm execution stage. The LLM agents can eventually accomplish the overall algorithm design by solving the subtasks stage-by-stage. 

Moreover, as shown by empirical evidence~\cite{RAG-origin, RAG-ssurvey}, supplementary domain knowledge can boost response accuracy and alleviate hallucinations in LLMs. Therefore, we provide intermediate real-time feedback information (e.g., for the chip placement task, we provide the current wirelength, the current overflow, a metric that measures the overall overlapping, the change of these two metrics made in this stage, as well as other layout statistics) to the LLM agents every time a stage is finished, and in this way, the LLM agents can be aware of how the target optimization problem is being solved by their algorithm design, update their domain knowledge, and improve their design in the right direction. Note that integrating domain knowledge in the process of utilizing data has been a promising direction (e.g., abductive learning~\cite{zhou2021abductive}) in artificial intelligence.

The Stagewise Design paradigm can significantly reduce the complexity of the tasks that LLM agents need to address due to the multi-stage decomposition. For example, in the chip placement task, it is difficult for LLMs to directly design an outstanding algorithm which can find a layout with non-overlapping cells and the minimum wirelength, while decomposing the difficult overall task into subtasks (e.g., finding a layout with fewer overlaps in the early stages and further optimizing it on wirelength in the late stages) can mitigate the difficulty of designing algorithms for subtasks, facilitating the generation of higher-quality algorithms by LLMs. Furthermore, the Stagewise Design paradigm enables the LLM agents to timely verify whether the algorithm design at the current stage meets the expected performance, and the collected valuable information about the target optimization problem at the current stage can guide the design of subsequent stages. In practice, the specific design scheme for each stage is autonomously decided by the LLM agents via analytical reasoning over the provided intermediate information.

\begin{figure}[t]
\centering
\includegraphics[width=0.8\textwidth]{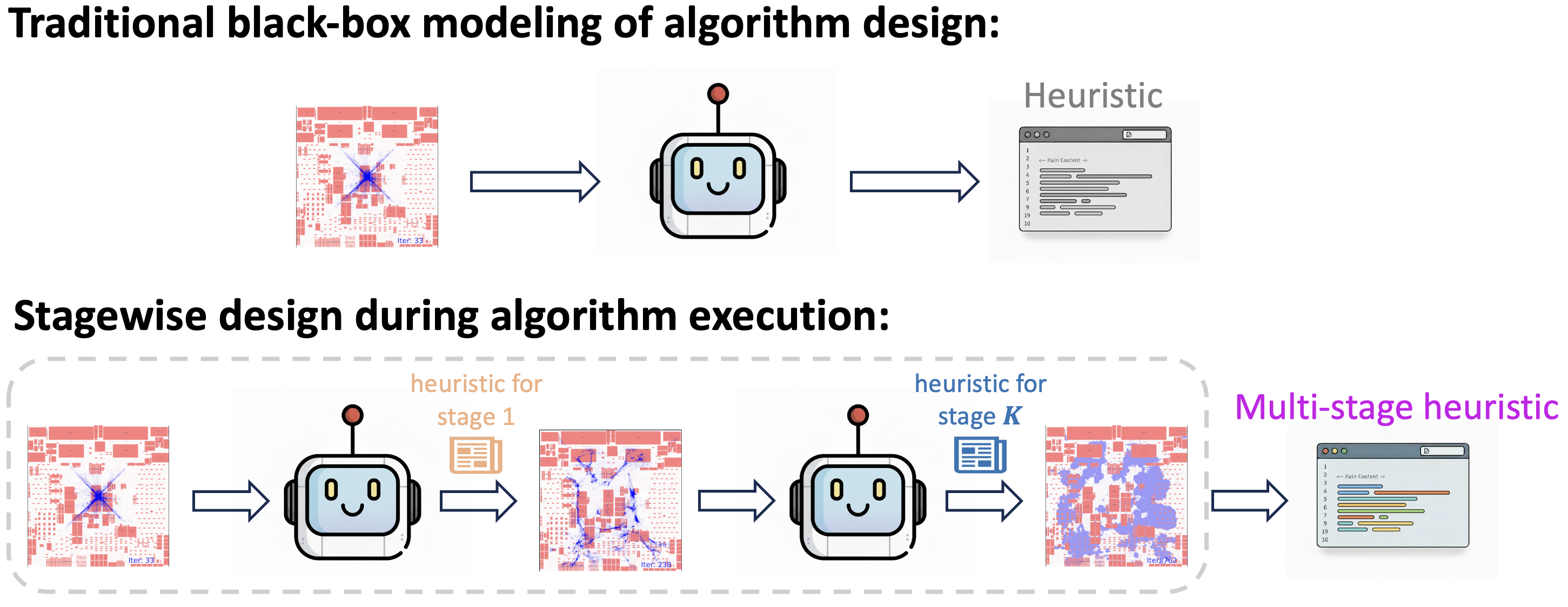}
\caption{The intuitive comparison between the traditional black-box modeling of algorithm design (upper) and the proposed Stagewise Design paradigm (lower).}
\label{plot:msh-comparison}
\end{figure}

The intuitive comparison between the proposed Stagewise Design and the traditional black-box modeling of algorithm design is shown in Figure~\ref{plot:msh-comparison}. The upper picture shows the traditional black-box modeling adopted by previous methods (i.e., FunSearch~\cite{FunSearch}, EoH~\cite{EoH}, ReEvo~\cite{ReEvo}, and AlphaEvolve~\cite{alphaevolve}), where LLMs generate a whole algorithm (or heuristic) directly and can only receive feedback after the whole evaluation of the designed algorithm, fully unaware of the optimization details of the target problem. Therefore, this modeling may misguide the optimization direction of LLMs and cause them to produce hallucinated algorithms. The lower picture shows the multi-stage designing process of Stagewise Design, where LLMs are grounded with real-time execution feedback to update their domain-specific understanding of the target problem and mitigate hallucinations, correcting the possible wrong optimization direction and producing algorithms with high-quality.

Formally, after the multi-stage decomposition, we have a sequence of goals $(g_1, g_2, \dots, g_K)$ for the subtask of each stage, where each goal $g_i$ reflects on the intermediate information $(I_0, I_1, \dots, I_{i-1})$ obtained so far, and $I_0$ represents the initial information about the task. The automatic way to derive the sequence of goals will be introduced in Section~\ref{MAS}. Given the goal $g_i$ for the subtask of each stage, the LLMs will derive a partial algorithm for the subtask of this stage: $A_i=\text{LLMs}(g_i),i=1,2,\dots K$, and the overall algorithm is derived as $\boldsymbol{A}=(A_1, A_2, \dots A_K)$.

\subsection{Multi-agent system}\label{MAS}

\begin{figure}[t]
\centering
\includegraphics[width=\textwidth]{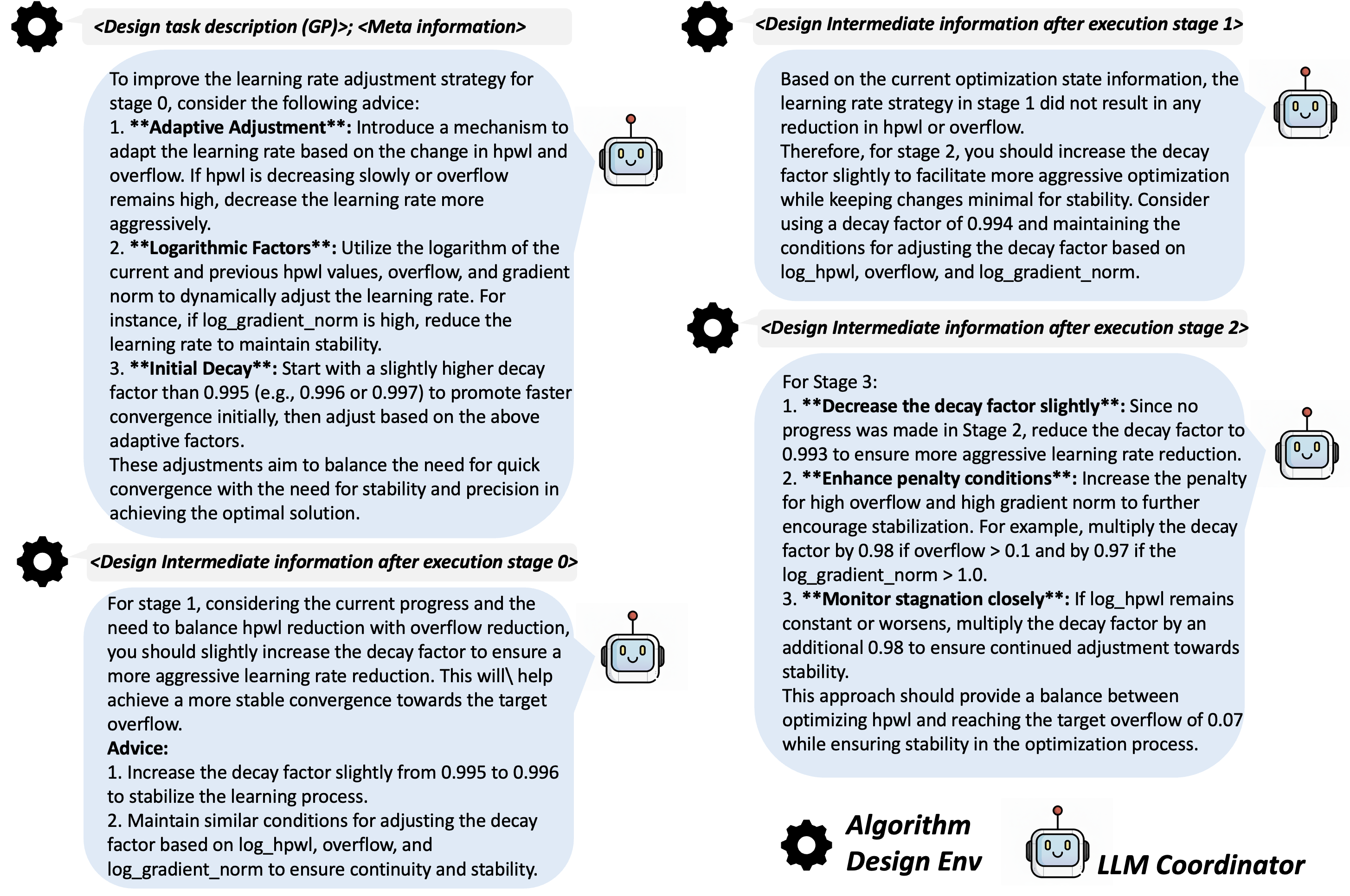}
\caption{One detailed example of the reflection reasoning of the coordinator between stages in the Stagewise Design paradigm.}
\label{plot:reflection-example}
\end{figure}

In order to tap into the full potential of LLMs in the proposed paradigm, we construct a multi-agent system, which facilitates the coordination of different agents to specialize in separate tasks of designing different algorithm components. In this multi-agent system, we have one individual LLM coder agent in charge of designing one specific algorithm component, whereas a dedicated coordinator agent performs reflection~\cite{reflecxion} on the current stage information and provides design guidelines for subsequent stages. For example, in the global placement (GP) procedure of chip placement, the rough layout of the cells is optimized by gradient descent, and we need to dynamically configure some parameters  (e.g., the learning rate, optimization step for each Lagrangian-relaxed problem, etc) throughout the algorithm execution. Therefore, we assign the task of designing the learning rate schedule to one coder agent and the task of designing the optimization step schedule to another coder agent, as well as the task of reflecting on the current stage and offering guidance to a coordinator agent.

The motivation of having multiple LLMs in charge of designing specific algorithm components is that empirically, when LLMs are required to generate code of diverse types and longer total lengths, they are more likely to produce code with more grammatical errors. Moreover, the multi-agent paradigm significantly reduces the complexity of the algorithm design space for individual LLM agents, enabling them to focus more on optimizing specific algorithm components and generating higher-quality designs. A dedicated coordinator agent plays an even more important role, reflecting on the current stage information and sending design guidelines to other LLM agents responsible for specific algorithm components for subsequent stages. This mechanism ensures all designing agents to achieve a consensus on the current stage information and align on the optimization direction for subsequent stages. Furthermore, this mechanism allows us to set different temperature parameters for different LLM agents, e.g., we can set a higher temperature for the coordinator agent because it needs to be more creative to reflect and give guidelines, while setting a lower temperature for designing agents because they need to produce legal codes. 

The reflection process of the coordinator is shown in Figure~\ref{plot:reflection-example}. When receiving the intermediate information about the historical execution stages through autonomous interaction with the algorithm design environment, the coordinator agent is prompted to perform reflection on the optimization effects of the partial algorithm for the previous stages and further offer guidance on the designing of the partial algorithm for the next stage, which is used to prompt the coder agents to produce the corresponding algorithm components.

Formally, for each stage $i$, the coordinator reflects on the execution information $(I_0, I_1, \dots, I_{i-1})$ obtained so far and generates the goal $g_i$ for the current stage, i.e., $g_i=\text{Coordinator}(I_0, I_1, \dots, I_{i-1})$, where $I_0$ is the initial information about the task. For the coder $j$, we assign it the design task of a separate algorithm component $\boldsymbol{C}_{j}$, and the coder $j$ generates the algorithm component at the current stage $i$ according to the current goal $g_i$: $C_{j,i}=\text{Coder}_j(g_i)$. 

\subsection{Global-local perspective mechanism}\label{operator}

Although multi-stage decomposition can greatly enhance the algorithm design capabilities of LLM agents, excessive focus on subtask optimization can occasionally lead to convergence to local optima, i.e., greedily optimizing partial algorithms $A_i, i=1,2,\dots,K$ by targeting the sequence of goals $(g_1, g_2, \dots, g_K)$ may not come up with a global optimal algorithm $\boldsymbol{A}^* = \mathop{\text{arg max}}_{\boldsymbol{A}\in \mathcal{A}} \text{score}(\boldsymbol{A})$. To avoid this issue, we introduce a "Global-Local Perspective" mechanism, which regards the Stagewise Design paradigm as an algorithm design operator derived from a local perspective, and further incorporates a global perspective to conduct a holistic analysis of some entire multi-stage algorithm designs (e.g., analyzing a selected multi-stage algorithm reference to get a basic understanding of the goals and the outcomes in each stage) and directly generate a new multi-stage algorithm without solving the subtasks stage-by-stage. Formally, the global perspective operator $\text{Op}_\text{global}$ is used to prompt the LLMs to produce a new overall multi-stage algorithm $\boldsymbol{A}_\text{new}=(A_{\text{new},1}, A_{\text{new},2}, \dots A_{\text{new},K})$ in one shot from selected multi-stage algorithm references, i.e., $\boldsymbol{A}_\text{new}=\text{LLMs}(\text{Op}_\text{global},\{\boldsymbol{A}^i_\text{ref}, \boldsymbol{I}^i_\text{ref}\}^k_{i=1})$, where $\boldsymbol{A}^i_\text{ref}=(A^i_{\text{ref},1}, A^i_{\text{ref},2}, \dots A^i_{\text{ref},K})$ is a multi-stage algorithm reference, $\boldsymbol{I}^i_\text{ref}=(I^i_{\text{ref},1}, I^i_{\text{ref},2}, \dots I^i_{\text{ref},K})$ is the information about $\boldsymbol{A}^i_\text{ref}$, and $k$ is the number of selected references. In specific, we design two types of global perspective operators (i.e., Global-Explore operator and Global-Enhance operator), which are introduced as follows, together with the local perspective operator Stagewise-Design. 

\begin{figure}[t]
\centering
\includegraphics[width=\textwidth]{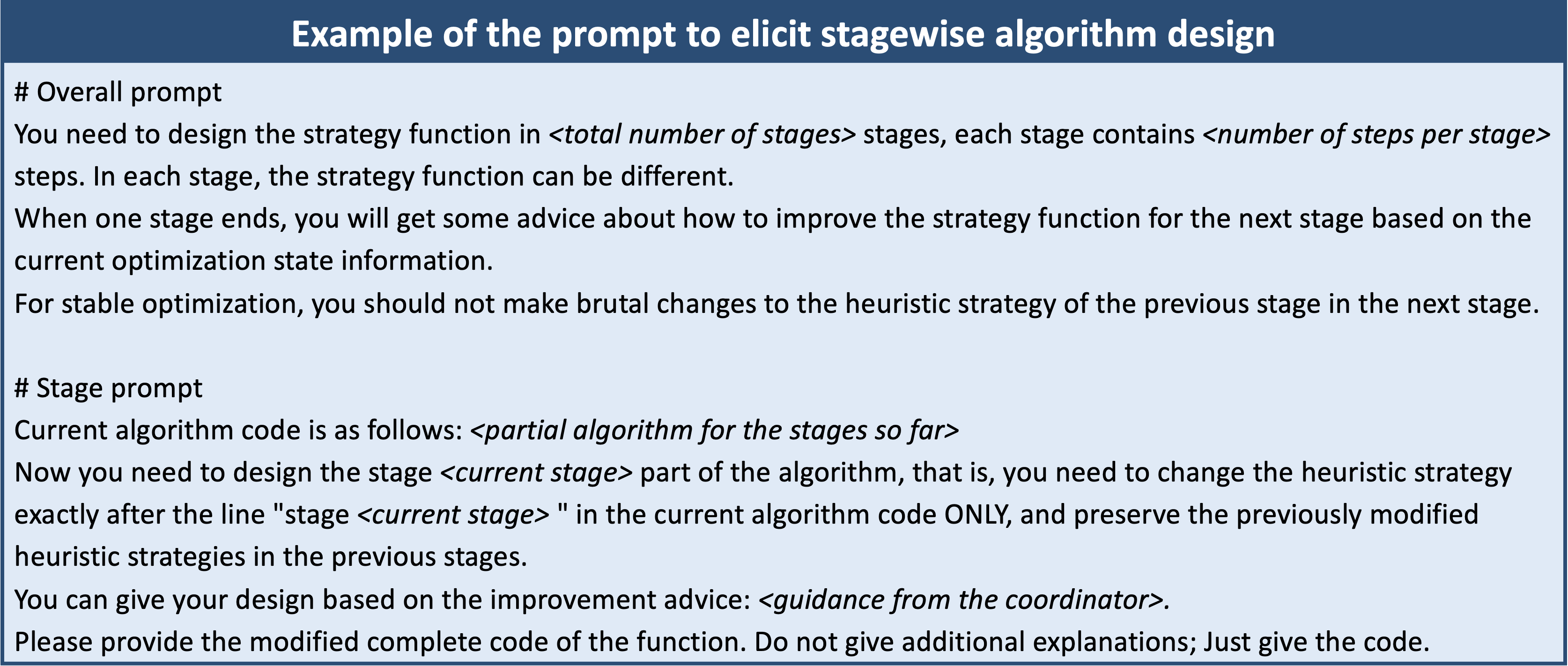}
\caption{The example prompt of the local perspective operator Stagewise-Design.}
\label{plot:prompt-example}
\end{figure}

\begin{figure}[t]
\centering
    \begin{subfigure}[b]{\textwidth}
    \centering
    \includegraphics[width=\textwidth]{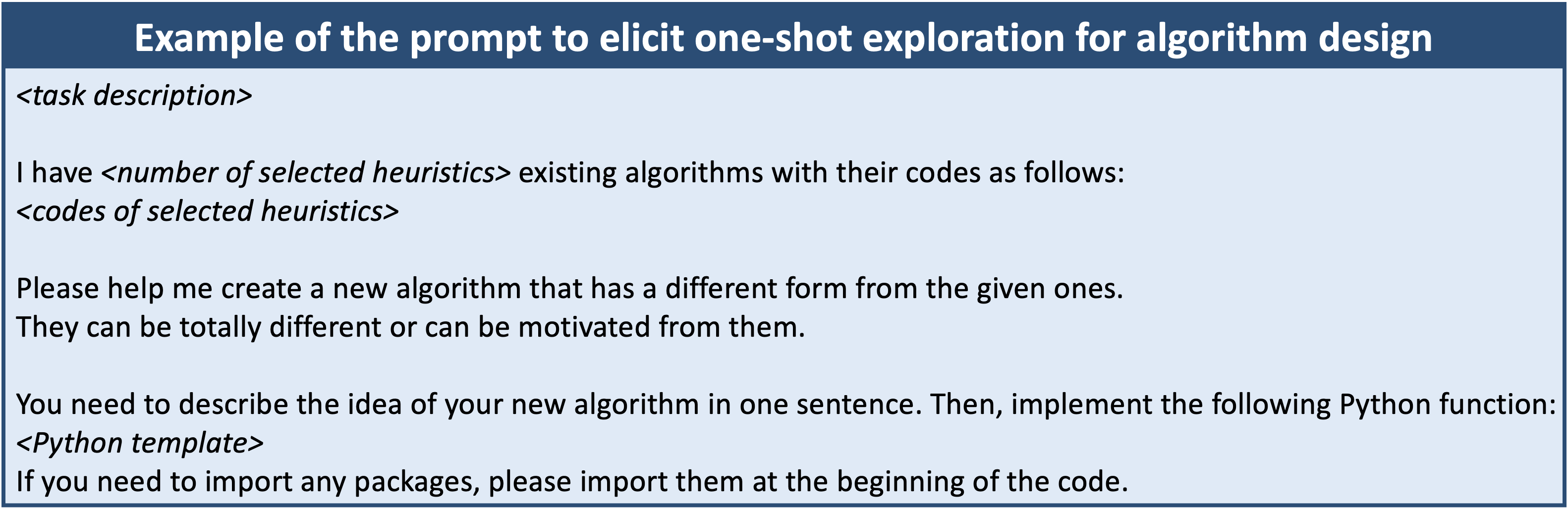} 
    \subcaption{Global-Explore operator}
    \end{subfigure}
    \begin{subfigure}[b]{\textwidth}
    \centering
    \includegraphics[width=\textwidth]{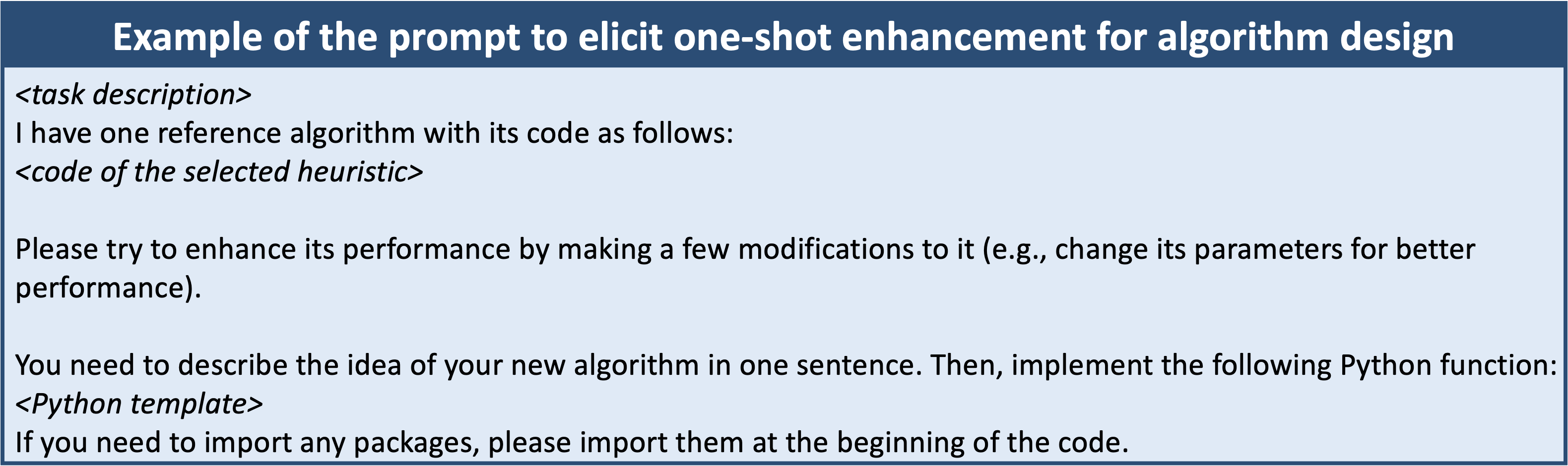}
    \subcaption{Global-Enhance operator}
    \end{subfigure}
\caption{The example prompt of the two global perspective operators: Global-Explore and Global-Enhance.}
\label{plot:global}
\end{figure}

\paragraph{Local perspective operator: Stagewise-Design} As introduced in Section~\ref{sec-MSH}, Stagewise Design decomposes the difficult algorithm design task into stages of subtasks, greatly enhancing the algorithm design capabilities of LLM agents. By this operator, the LLMs are prompted to design algorithm components stage-by-stage; they are also provided intermediate statistics to better reflect on the stage information and then make better designs for the next stage accordingly. The example prompt of the local perspective operator is demonstrated in Figure~\ref{plot:prompt-example}.

\paragraph{Global perspective operators: Global-Explore and Global-Enhance} In the Global-Explore operator, $k$ previous multi-stage algorithm designs are first sampled from the multi-stage heuristic population (by applying the selection strategy introduced in Section~\ref{sec-framework}) as the reference designs, and then we prompt the LLMs to make a new design that has a different form (in the sense of ideas) from the given ones. The new design can be motivated or totally different from the reference designs. This operator can prevent the LLMs from consistently producing similar algorithm designs without any significant improvement, as well as increasing the probability of finding the optimal design backbone. In the Global-Enhance operator, one design is first sampled from the population, and then we prompt the LLMs to enhance its performance by making a few modifications to it (e.g., changing its parameters for better performance). This operator encourages the LLMs to exploit the current design for better performance. Note that we do not necessarily sample the current best design (as the selection strategy introduced in Section~\ref{sec-framework}) for LLMs to enhance because some underperformed designs may still have large potential for excellent performance, and the top performance of the current best design may indicate less space for further improvement. The example prompts of the two introduced global perspective operators are demonstrated in Figure~\ref{plot:global}.

In our implementation, the above three operators (i.e., Stagewise-Design, Global-Explore, and Global-Enhance) are applied sequentially with a frequency ratio of 1:1:1, helping the LLMs to balance the local optimization of subtasks and the global optimization of the overall algorithm design, as well as balancing the quality and diversity of the generated algorithms. 

\subsection{Comparison to previous automated algorithm design methods with LLMs}

Previous LLM-based methods~\cite{FunSearch, alphaevolve, EoH, ReEvo} typically adopt the evolutionary framework to help LLMs iteratively improve the algorithm designs by focusing on important information and promising designs with selected references, but in a black-box modeling, where LLMs are unaware of the details about how exactly the designed algorithms take effect, except for a final score. In contrast, EvoStage decomposes the algorithm design task into multiple stages and assists the LLMs to solve the task stage-by-stage with intermediate feedback (as shown in Figure~\ref{plot:msh-comparison}), thereby significantly reducing the complexity of the tasks that LLM agents need to address at one time. Moreover, through reflecting on the intermediate feedback during execution, LLMs are grounded in the actual optimization effects produced by their designed algorithms, which enables the LLMs to timely update their domain knowledge without the need for similar examples to refer to, and thus significantly enhances their capabilities in scenarios where similar examples are scarce and instances exhibit substantial discrepancies. 

To cut down the design space of the algorithm, we construct a multi-agent system where each LLM coder agent is assigned the design task of an algorithm component, and an LLM coordinator agent is responsible for reflecting on the intermediate information and offering guidance to different LLM coder agents, which keeps the consistency of the overall design. Furthermore, we design two types of operators (i.e., the global perspective operator and the local perspective operator) to design multi-stage heuristics in one shot from the global perspective, as well as stage-by-stage from the local perspective, helping escape local optima in the process of algorithm design.

These three novel components make our proposed EvoStage significantly distinct from the previous LLM-based algorithm design methods~\cite{FunSearch, alphaevolve, EoH, ReEvo}, which better align with the capacities of LLMs and balance exploration and exploitation well during optimization. Thus, EvoStage can improve the efficiency of the overall algorithm design procedure and derive algorithm designs with higher quality within limited budgets, enabling automated algorithm design even in industrial scenarios with limited evaluation budgets and few high-quality samples, e.g., chip placement, which will be shown in Section~\ref{chip-placement}.

\section{Applying EvoStage to designing parameter configuration schedules of Adam optimizer for chip placement} \label{chip-placement}

In this section, we demonstrate the excellent performance of the proposed EvoStage for automated algorithm design in industry-level scenarios. In particular, we first apply EvoStage to design schedules of two key parameters (learning rate and optimization step) for the popular gradient-based optimizer Adam in the optimization of global placement of chips; then, we further apply EvoStage to design the learning rate schedules as well as the density weight schedules of a commercial-grade 3D chip placement tool on a real-world industrial 3D chip case.

\subsection{Global placement}\label{sec-GP}
\paragraph{Global placement} Global placement (GP) is a critical step of the overall chip placement procedure~\cite{chu2009placement}, which has a direct and
significant impact on the performance, power consumption, and area (PPA) of the final chip design, as it determines the locations of the cells within the given chip layout area~\cite{markov2012progress}. As aiming to provide a high-quality rough overall layout of all the cells, GP is often required to achieve the target overflow (a metric that measures the overall overlapping of a chip) with the lowest half-perimeter wirelength (HPWL), as well as being as fast as possible~\cite{lin2020dreamplace}. As for formulation, GP is typically formulated as a constrained optimization problem with the HPWL as the objective and the density as the constraint: 
\begin{equation}
\begin{gathered}
    \underset{\textbf{x}, \textbf{y}}{\min}\text{ HPWL}(\textbf{x},\textbf{y})=\underset{\textbf{x}, \textbf{y}}{\min}\sum\nolimits_{e\in E}\text{HPWL}_e(\textbf{x},\textbf{y}),\\
\text{s.t. } \text{D}(\textbf{x}, \textbf{y}) \le d_t,
\end{gathered}
\label{eq-gp-constrained}
\end{equation}
where $(\textbf{x}, \textbf{y})$ denotes the 2-D positions of all the cells, $E$ denotes the set of nets,  $\text{HPWL}_e(\textbf{x},\textbf{y})= (\text{max}_{i\in e} x_i - \text{min}_{i\in e} x_i) + (\text{max}_{i\in e} y_i - \text{min}_{i\in e} y_i)$ denotes the HPWL of each net $e$, $\text{D}(\textbf{x}, \textbf{y})$ denotes the density of each location of the layout, and $d_t$ denotes the target density. Note that a net encompasses multiple cells and signifies their interconnectivity during the routing phase.

\paragraph{Existing GP methods} Analytical placement~\cite{essential-issues-in-analytical} is one type of state-of-the-art GP methods, among which nonlinear placement~\cite{chen2008ntuplace3, lu2015eplace, cheng2018replace} achieves the best results. Nonlinear placement typically solves the constrained optimization problem of GP in Eq.~(\ref{eq-gp-constrained}) by the penalty method, i.e., solving a sequence of Lagrangian-relaxation problems with an increasing Lagrangian multiplier, which are formulated as follows:
\begin{equation}
    \underset{\textbf{x}, \textbf{y}}{\text{min}}\;\left(\sum\nolimits_{e\in E} \text{WL}_e(\textbf{x},\textbf{y})\right) + \lambda \cdot  \text{D}(\textbf{x}, \textbf{y}),
    \label{target}
\end{equation}
where $\text{WL}_e(\textbf{x},\textbf{y})$ is a smooth approximation of $\text{HPWL}_e(\textbf{x},\textbf{y})$, and $\lambda$ is the Lagrangian multiplier (also called density weight), which controls the importance of the density penalty function $\text{D}(\textbf{x}, \textbf{y})$. For a smoother optimization, ePlace~\cite{lu2015eplace} and RePlace~\cite{cheng2018replace} compute a density penalty function analogous to the potential energy of an electrostatic system (where cells are modeled as charges, and the density gradient is modeled as the electric field), which is smoother than the original density function and has a good correlation with it. 

Despite the outstanding performance, the nonlinear methods are typically time-consuming, and thus cannot meet the fast verification requirement of industrial scenarios, especially for large-scale chip cases~\cite{lin2020dreamplace}. Therefore, accelerating algorithms like DREAMPlace~\cite{lin2020dreamplace} and Xplace~\cite{xplace} are proposed to accelerate the optimization process of gradient descent with modern GPUs and deep learning toolkits. Specifically, DREAMPlace~\cite{lin2020dreamplace} integrates modern gradient descent optimizers (e.g., Adam, Nesterov, etc) and adapts neural network training techniques to GP tasks, which achieves superior performance, representing the most widely adopted and advanced open-source placer at present.

\paragraph{Settings of EvoStage} In this experiment, we choose Adam~\cite{kingma2014adam} as the optimizer of DREAMPlace~\cite{lin2020dreamplace} for the gradient descent optimization process, and apply our proposed EvoStage to design the learning rate schedule for the Adam optimizer, which, as shown empirically, plays a very crucial role in the neural network training task, as well as the GP task. Meanwhile, we apply EvoStage to design the optimization step schedule for the sequence of Lagrangian-relaxation problems, analogous to the number of epochs in the neural network training task, which has a significant impact on the final HPWL, analogous to the training loss in the neural network training task. Empirically, using more optimization steps for a certain Lagrangian-relaxation problem can lead to more optimal results with respect to this Lagrangian-relaxation problem, but at the same time, it is probably easier to fall into local optima and is more time-consuming. It is worth noting that, to demonstrate the effectiveness of our method, for the parameters that are not concerned in the algorithm design process of our method, we substitute simple and easy configuration strategies for the complicated expert-designed strategies that have been iteratively improved in previous placement engines~\cite{lu2015eplace, cheng2018replace, lin2020dreamplace}.

For the evolution process, we maintain a population of 5 algorithm individuals, reproduce 5 individuals for one generation, and make an evolution for a total of 5 generations (i.e., a total of 25 real evaluations). In this setting, an algorithm individual consists of two multi-stage heuristics, corresponding to the two designed algorithm components, i.e., the learning rate schedule and the optimization step schedule. We set the number of stages as 4 for the Stagewise-Design operator. The implementation details can be found in Appendix~\ref{implementation-detail-GP}.

\begin{table*}[t!]
    \caption{GP HPWL results ($\times 10^6$) of the ISPD 2005 benchmark and the ICCAD 2015 benchmark achieved by two state-of-the-art GP placers (DREAMPlace-Nesterov~\cite{lin2020dreamplace} and Xplace-NN~\cite{xplace}) and our method EvoStage. Results in bold denote the current best results (the historically state-of-the-art results, to the best of our knowledge) that surpass the previous best results ever achieved. The average ratio takes the results of the proposed EvoStage as baselines.}
    \label{tab:goat}
    \centering
    \resizebox{\linewidth}{!}{
    \begin{tabular}{c|ccccccccc}
    \toprule
     Chip case& adaptec1 & adaptec2 & adaptec3 & adaptec4 & bigblue1 & bigblue2 & bigblue3 & bigblue4 & Avg \\
     \midrule
    DREAMPlace-Nesterov~\cite{lin2020dreamplace} &   70.28  &   79.23  &   185.74  &   168.81  & 87.36 &   131.10  &   291.62  &   725.47  &  1.009\\
    
    Xplace-NN~\cite{xplace} & 71.34 & 80.35 & 195.62 & 175.53 &  87.34 & 135.53 & 296.01 & 734.29 & 1.033\\

    EvoStage (ours) &   \textbf{ 69.76 } &   \textbf{ 78.59 } &   \textbf{ 183.50 } &   \textbf{ 166.67 } &   \textbf{ 86.76 } &   \textbf{ 130.75}  &   \textbf{ 286.23 } &   \textbf{ 720.69 } &   \textbf{ 1.000 }\\
    \midrule
    Chip case & superblue1 & superblue3 & superblue4 & superblue5 & superblue7 & superblue10 & superblue16 & superblue18 & Avg\\
    \midrule
    DREAMPlace-Nesterov~\cite{lin2020dreamplace} & 390.49 &  441.02 & 291.23 & 460.48 & 553.56 & 868.69 &  403.17 & 222.36 & 1.019\\

    Xplace-NN~\cite{xplace} &   388.12  & 444.84 &   289.76  &   456.79  &   548.96  &   867.67  & 408.34 &   220.34  &  1.017\\

    EvoStage (ours) &   \textbf{ 381.98 } &   \textbf{ 433.81 } &   \textbf{ 287.01 } &   \textbf{ 448.32 } &   \textbf{ 544.00 } &   \textbf{ 852.83 } &   \textbf{ 399.98 } &   \textbf{ 216.89 } &   \textbf{ 1.000 }\\
    \bottomrule
    \end{tabular}
    }
\end{table*}

\paragraph{Open-source chip benchmarks} 
We conduct experiments on two popular open-source benchmarks for GP: ISPD 2005~\cite{ispd2005} and ICCAD 2015~\cite{iccad2015}, with the stop overflow set as 0.07 for the ISPD 2005 benchmark and 0.1 for the ICCAD 2015 benchmark. For the ISPD 2005 benchmark, we provide no design examples to fully demonstrate the ability of the compared methods, while for the ICCAD 2015 benchmark, we provide one example of a simple exponential decay schedule of learning rate due to the difficulty of this benchmark.

\paragraph{Comparison with state-of-the-art GP methods} Table~\ref{tab:goat} demonstrates GP HPWL results (the smaller, the better) achieved by our method EvoStage and two placers that are elaborately designed and fine-tuned by human experts. The upper 8 chip cases belong to the ISPD 2005 benchmark, while the lower 8 cases belong to the ICCAD 2015 benchmark. As introduced before, EvoStage simultaneously designs the learning rate schedule and the optimization step schedule for the sequence of the Lagrangian-relaxation problems. DREAMPlace-Nesterov~\cite{lin2020dreamplace} denotes the fine-tuned DREAMPlace version with Nesterov optimizer and an accelerated line search mechanism to determine the maximum steplength; Xplace-NN~\cite{xplace} denotes the Xplace version with a neural network enhancement for density gradient prediction. To the best of our knowledge, these two placers are the state-of-the-art open-source placers to date and have yielded state-of-the-art GP performance on widely used open-source datasets. The results in bold in Table~\ref{tab:goat} denote the current best results that surpass the previous best results ever achieved (shown in~\cite{lin2020dreamplace} and~\cite{xplace}). To the best of our knowledge, our method achieves the historically state-of-the-art GP HPWL results on every chip case of the two open-source benchmarks within just 25 evaluations. We provide some examples of the best-performing generated algorithms that exhibit adaptive adjustment capabilities distinct from those designed by human experts in Appendix~\ref{generated-algorithm} for a more intuitive understanding. We also provide some examples of the layout plots in Appendix~\ref{layout-plots} for a more intuitive layout comparison.

\begin{table*}[t!]
    \caption{GP HPWL results ($\times 10^6$) of the ISPD 2005 benchmark. For all LLM-based methods, only the learning rate schedule is designed (i.e., for EoH, AlphaEvolve, and EvoStage-single), except for EvoStage (it designs both the learning rate and the optimization step schedule); a maximum budget of 25 evaluations is used. Pass Rate measures the basic quality of the codes generated by LLM-based methods. One designed code is counted as passed if it is legal and achieves the target overflow with reasonable HPWL (i.e., less than 1e9). The average ratio takes the DREAMPlace-Adam's results as baselines. The results in bold denote the best results when only designing the learning rate schedule, and the results in bold with underline denote the better results achieved by designing both the learning rate schedule and the optimization step schedule.}
    \label{tab:ispd}
    \centering
    \resizebox{\linewidth}{!}{
    \begin{tabular}{c|c|cc|cc|cc|cc}
    \toprule
     \multirow{2}{*}{Chip case}&\multicolumn{1}{c}{DREAMPlace-Adam~\cite{lin2020dreamplace}}& \multicolumn{2}{c}{EoH~\cite{EoH}} &\multicolumn{2}{c}{AlphaEvolve~\cite{alphaevolve}} & \multicolumn{2}{c}{EvoStage-single (ours)} &\multicolumn{2}{c}{EvoStage (ours)}\\
    & HPWL  & HPWL & Pass Rate (\%) & HPWL & Pass Rate (\%) & HPWL & Pass Rate (\%) & HPWL & Pass Rate (\%)\\
    \midrule
    adaptec1 & 73.27 & 75.84 & 20.00 & 73.46 & 24.00 & \textbf{70.40} & \textbf{60.00}& \textbf{\underline{69.76}} & \textbf{\underline{80.00}} \\
    adaptec2 & 82.28 & 148.00 & 8.00 & 87.78 & 16.00 & \textbf{80.17} & \textbf{64.00}&\textbf{\underline{78.59}} & \textbf{\underline{80.00}}\\
    adaptec3 & 189.17 & 189.44 & 4.00 & 187.78 & 20.00 & \textbf{186.70} & \textbf{64.00}& \textbf{\underline{183.50}} & \textbf{\underline{68.00}}\\
    adaptec4 & 172.15 & 169.83 & 36.00 & 173.38 & 16.00 & \textbf{168.50} & \textbf{64.00}&\textbf{\underline{166.67}} & \textbf{\underline{72.00}}\\
    bigblue1 & 88.93 & 92.64 & 44.00 & 95.45 & 32.00 & \textbf{87.60} & \textbf{100.00}&\textbf{\underline{86.76}} & 88.00\\
    bigblue2 & 133.26 & 132.29 & 60.00 & 135.49 & 64.00 & \textbf{131.85} & \textbf{68.00}&\textbf{\underline{130.75}} & \textbf{\underline{76.00}}\\
    bigblue3 & 305.24 & 515.29 & 4.00 & 321.20 & 20.00 & \textbf{294.86} & \textbf{64.00}&\textbf{\underline{286.23}} & \textbf{\underline{68.00}}\\
    bigblue4 & 737.65 & 939.16 & 8.00 & 755.05 & 10.00& \textbf{732.05} & \textbf{52.00}& \textbf{\underline{720.69}} & \textbf{\underline{92.00}}\\
    \midrule
    Avg & 1.000 & 1.227 & 23.00 & 1.029 & 25.25 & \textbf{0.979} & \textbf{67.00}& \textbf{\underline{0.964}} & \textbf{\underline{78.00}}\\
    \bottomrule
    \end{tabular}
    }
\end{table*}

\begin{table*}[t!]
    \caption{GP HPWL results ($\times 10^6$) of the ICCAD 2015 benchmark. For all LLM-based methods, only the learning rate schedule is designed (i.e., for EoH, AlphaEvolve, and EvoStage-single), except for EvoStage (it designs both the learning rate and the optimization step schedule); a maximum budget of 25 evaluations is used. Pass Rate measures the basic quality of the codes generated by LLM-based methods. One designed code is counted as passed if it is legal and achieves the target overflow with reasonable HPWL (i.e., less than 1e9). The average ratio takes the DREAMPlace-Adam's results as baselines. The results in bold denote the best results when only designing the learning rate schedule, and the results in bold with underline denote the better results achieved by designing both the learning rate schedule and the optimization step schedule.}
    \label{tab:iccad}
    \centering
    \resizebox{\linewidth}{!}{
    \begin{tabular}{c|c|cc|cc|cc||cc}
    \toprule
     \multirow{2}{*}{Chip case}&\multicolumn{1}{c}{DREAMPlace-Adam~\cite{lin2020dreamplace}}& \multicolumn{2}{c}{EoH~\cite{EoH}} &\multicolumn{2}{c}{AlphaEvolve~\cite{alphaevolve}} & \multicolumn{2}{c}{EvoStage-single (ours)} & \multicolumn{2}{c}{EvoStage (ours)}\\
    & HPWL  & HPWL & Pass Rate (\%) & HPWL & Pass Rate (\%) & HPWL & Pass Rate (\%) & HPWL & Pass Rate (\%)\\
    \midrule
    superblue1 & 390.08 & 386.45 & 84.00 & 388.87 & 48.00 & \textbf{386.13} & \textbf{100.00} & \textbf{\underline{381.98}} & \textbf{\underline{100.00}}\\
    superblue3 & 462.22 & 582.64 & 4.00 & 462.31 & 4.00 & \textbf{449.44} & \textbf{68.00} & \textbf{\underline{433.81}} & \textbf{\underline{84.00}}\\
    superblue4 & 292.11 & 289.62 & 80.00 & 291.62 & 44.00 & \textbf{289.13} & \textbf{88.00} & \textbf{\underline{287.01}} & 80.00\\
    superblue5 & 452.34 & 453.57 & 28.00 & 452.31 & 16.00 & \textbf{451.16} & \textbf{84.00} & \textbf{\underline{448.32}} & \textbf{\underline{92.00}}\\
    superblue7 & 556.69 & 554.63 & 24.00 & 554.18 & 32.00 & \textbf{549.40} & \textbf{72.00} & \textbf{\underline{544.00}} & \textbf{\underline{84.00}}\\
    superblue10 & 861.29 & 859.82 & 60.00 & 860.51 & 32.00 & \textbf{856.25} & \textbf{80.00} & \textbf{\underline{852.83}} & \textbf{\underline{88.00}}\\
    superblue16 & 402.58 & 401.12 & 60.00 & 401.38 & 16.00 & \textbf{400.11} & \textbf{80.00} & \textbf{\underline{399.98}} & \textbf{\underline{84.00}}\\
    superblue18 & 224.86 & 224.48 & 28.00 & 223.81 & 24.00 & \textbf{223.40} & \textbf{84.00} & \textbf{\underline{216.89}} & \textbf{\underline{100.00}}\\
    \midrule
    Avg & 1.000 & 1.029 & 46.00 & 0.998 & 27.00 & \textbf{0.990} & \textbf{82.00} & \textbf{\underline{0.977}} & \textbf{\underline{89.00}}\\
    \bottomrule
    \end{tabular}
    }
\end{table*}

\paragraph{Comparison with LLM-based algorithm design methods} Furthermore, we compare our method EvoStage to two well-established LLM-based algorithm design methods: EoH~\cite{EoH} and AlphaEvolve~\cite{alphaevolve}. For a fair comparison, we report the GP HPWL results (as shown in Tables~\ref{tab:ispd} and~\ref{tab:iccad}) obtained by all the methods, designing only the learning rate schedule and using a budget of only 25 evaluations. Note that the version of EvoStage designing only the learning rate schedule is called EvoStage-single. DREAMPlace-Adam~\cite{lin2020dreamplace} denotes the DREAMPlace version using the Adam optimizer, serving as a baseline. For AlphaEvolve, we use an open-source version named "OpenEvolve"\footnote{https://github.com/algorithmicsuperintelligence/openevolve} since the original paper~\cite{alphaevolve} does not provide an official implementation. The "Pass Rate" metric measures the quality of the overall evolution process of the LLM-based methods. If one designed code is legal and achieves the target overflow with reasonable HPWL (i.e., less than 1e9), we count it as one passed designed code. The "Pass Rate" metric is calculated as the ratio of the passed designed code to the total number of generated codes. The average ratios and improvements of the different methods reported are based on the GP results of DREAMPlace-Adam. 

As shown in Tables~\ref{tab:ispd} and~\ref{tab:iccad}, when only designing the learning rate schedule, our method EvoStage-single achieves the best GP HPWL results and the highest Pass Rate compared to other LLM-based methods. This is expected because our proposed Stagewise Design paradigm better aligns with LLM's ability and helps them make full use of each evaluation for better overall designs. The highest Pass Rate also implies that the provided intermediate information during algorithm execution and the corresponding reflection help the LLM agents to get a better understanding of the target optimization problem and correctly adjust the optimization direction. While EoH~\cite{EoH} and AlphaEvolve~\cite{alphaevolve} struggle to come up with a better design compared to the expert-designed schedule DREAMPlace-Adam~\cite{lin2020dreamplace} due to their black-box modeling that deprives LLMs of any awareness of the target problem and disables the LLMs from utilizing additional information to correct possibly wrong design directions in a limited budget. Moreover, when simultaneously designing the learning rate schedule and the optimization step schedule for the sequence of Lagrangian-relaxed problems, our proposed EvoStage achieves even better GP results and Pass Rate, as shown in the last column of Tables~\ref{tab:ispd} and~\ref{tab:iccad}, which shows the effectiveness of the proposed multi-agent system. Note that the recent work~\cite{evoplace} has attempted to directly use LLMs to generate GP-related algorithms (e.g., the macro initialization algorithm, which determines the initial position of the macro cells on the chip layout) within the previous evolution framework, but their results are not comparable with ours, because the authors modified the ISPD2005 benchmark~\cite{ispd2005} to make the macros and I/O objects movable with their original shapes.

\begin{figure}[t]
\centering
    \includegraphics[width=\textwidth]{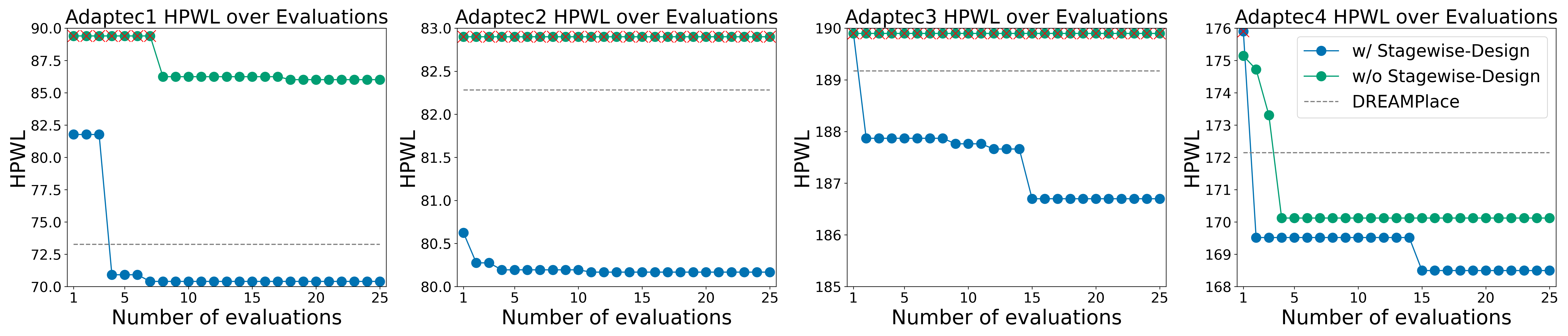} 
\caption{The HPWL comparison of EvoStage with or without the Stagewise-Design operator over evaluations on 4 chip cases, i.e., Adaptec1--Adaptec4. The red \textcolor{red}{$\times$} denotes a failed run of placement which does not achieve the target overflow.}
\label{fig:ablation}
\end{figure}

\paragraph{Ablation of EvoStage} To verify the effectiveness of our proposed Stagewise Design paradigm, we conduct ablation experiments to compare the performance of the complete version of EvoStage and that without the Stagewise-Design operator. The HPWL results over evaluations are shown in Figure~\ref{fig:ablation}, where a "$\times$" in an evaluation denotes that the design in that evaluation does not achieve the target overflow. To directly demonstrate the effectiveness of the Stagewise-Design operator, we apply the two compared versions of EvoStage to only design the learning rate schedules. As shown in Figure~\ref{fig:ablation}, when equipped with the Stagewise-Design operator, EvoStage can efficiently come up with a design with high performance, even just in a few shots; while without the Stagewise-Design operator, the performance of EvoStage degrades seriously, and the final design is worse than the expert-designed schedule DREAMPlace in three of four cases, analogous to the struggling performance of previous LLM-based algorithm design methods EoH~\cite{EoH} and AlphaEvolve~\cite{alphaevolve} as observed in Tables~\ref{tab:ispd} and~\ref{tab:iccad}. These results indicate that the Stagewise Design paradigm does really help the LLM agents make full use of the evaluation information and timely refine their optimization direction with a better understanding of the target problem, leading to fewer hallucinations and better designs.

\subsection{3D IC and 3D placement} 

\begin{figure}[t]
\centering
\includegraphics[width=0.3\textwidth]{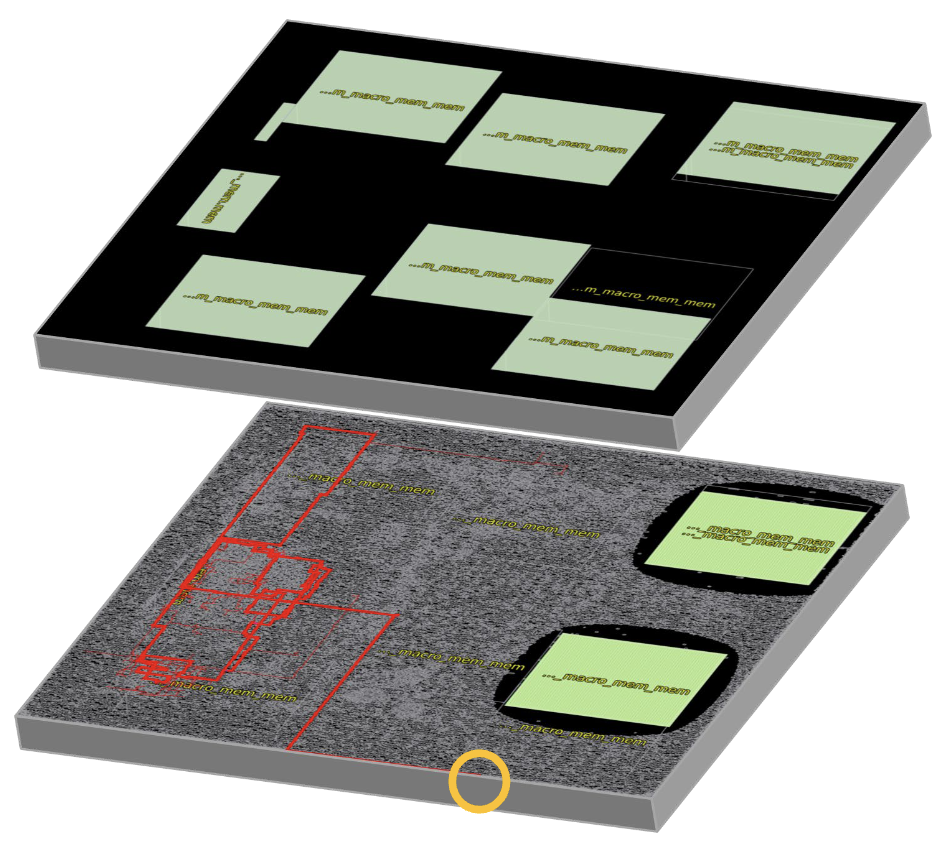}
\caption{An intuitive illustration of the 3D memory-on-logic stacking, where on the top die, the macro cells with large sizes (e.g., the memories) are placed (denoted in green), and the standard cells with smaller sizes (e.g., the logic gates) are placed on the bottom die (denoted in gray). Note that some macro cells densely interconnected with the standard cells are also allowed to be placed on the bottom die.
}
\label{plot:3d}
\end{figure}

\paragraph{3D global placement} As semiconductor manufacturing processes gradually approach their physical limits and Moore's Law gradually loses its validity, the area of electronic design automation (EDA) has shifted its focus from 2D chips to 3D chips to fabricate chips with higher integration density~\cite{shi2025open3dbench}. For example, AMD has proposed AMD 3D V-Cache Technology, and achieves more than 200x interconnect density compared to on-package 2D chiplet based on AMD internal data~\cite{amd3d}; Intel has also proposed its pioneering Lakefield~\cite{intel3d} utilizing its 3D stacking technology, Foveros. One of the most feasible approaches for 3D IC in real-world industrial scenarios is the two-die paradigm called "memory-on-logic", where the macro cells with large sizes (e.g., the memories) will be placed on the top die, while the standard cells with smaller sizes (e.g., the logic gates) will be placed on the bottom die. This is exactly the paradigm that commercial-grade 3D chip placement tools often adopt. Figure~\ref{plot:3d} gives an example illustration of the 3D memory-on-logic stacking, where some macro cells densely interconnected with the standard cells are also allowed to be placed on the bottom die. In this paradigm, the placement of the top and bottom dies is usually collaboratively optimized after the high-quality partition that allocates each cell to a specified die in advance. Specifically, the metrics like the HPWL value of the two dies (i.e., HPWL-logic and HPWL-mem) and the overflow value of the two dies (i.e., the overflow-logic and overflow-mem) need to be optimized.

\paragraph{Settings of EvoStage} We apply EvoStage to a commercial-grade 3D chip placement tool, which adopts the memory-on-logic 3D placement technology, to autonomously design a real-world industrial 3D chip case that is modified from a real-world industrial 2D chip design. The commercial-grade 3D chip placement tool is also accelerated by GPUs, where the cell placement is optimized through minimizing a group of elaborately designed losses (i.e., a joint wirelength loss and a joint density loss under the penalty method) by gradient descent. In order to improve the HPWL metrics while achieving the target overflow, as well as significantly improve the GP efficiency at the same time, EvoStage is applied to simultaneously design the learning rate schedule of the gradient optimizer Adam and the schedule of density weight (i.e., the Lagrangian multiplier of each Lagrangian-relaxation problem, which balances the optimization of wirelength and density). The population size, the number of generations for the evolution process, and the number of stages for the Stagewise-Design operator are set as same as those used in 2D GP in Section~\ref{sec-GP}, which equal to 5, 5, and 4, respectively.  

\begin{table*}[t!]
    \caption{Desensitized GP results (HPWL and overflow) of the placement of a real-world industrial 3D chip case. The base 3D placer is a commercial-grade 3D chip placement tool. To improve it, EvoStage simultaneously designs the learning rate schedule of the gradient optimizer Adam and the density weight (i.e., the Lagrangian multiplier of each Lagrangian-relaxation problem) schedule. HPWL-logic and HPWL-mem denote the half-perimeter wirelength of the logic die and the memory die, respectively; Overflow-logic and Overflow-mem measure the overlap issues of the logic die and the memory die, respectively; and Iteration denotes the total number of gradient descent optimization steps. Note that we desensitize the results by computing the ratios that take the base 3D placer's results as baselines.}
    \label{tab:3d}
    \centering
    \begin{tabular}{c|ccccc}
    \toprule
      & HPWL-logic & HPWL-mem & Overflow-logic & Overflow-mem & Iteration \\
    \midrule
     Base 3D placer & 1.00  & 1.00  & \textbf{1.00}  & 1.00  & 1.00  \\
     \midrule
     EvoStage & \textbf{ 0.91 } & \textbf{ 0.92 } & \textbf{1.00} & \textbf{ 0.99 } & \textbf{ 0.48 }  \\
     \midrule
     Improvement (\%) & 9.24 & 7.51 & 0.00 & 0.52 & 52.21 \\
    \bottomrule
    \end{tabular}
\end{table*}

\begin{figure}[t!]
    \centering
    \resizebox{\linewidth}{!}{
    \begin{subfigure}[b]{0.48\textwidth}
        \centering
        \includegraphics[width=\textwidth]{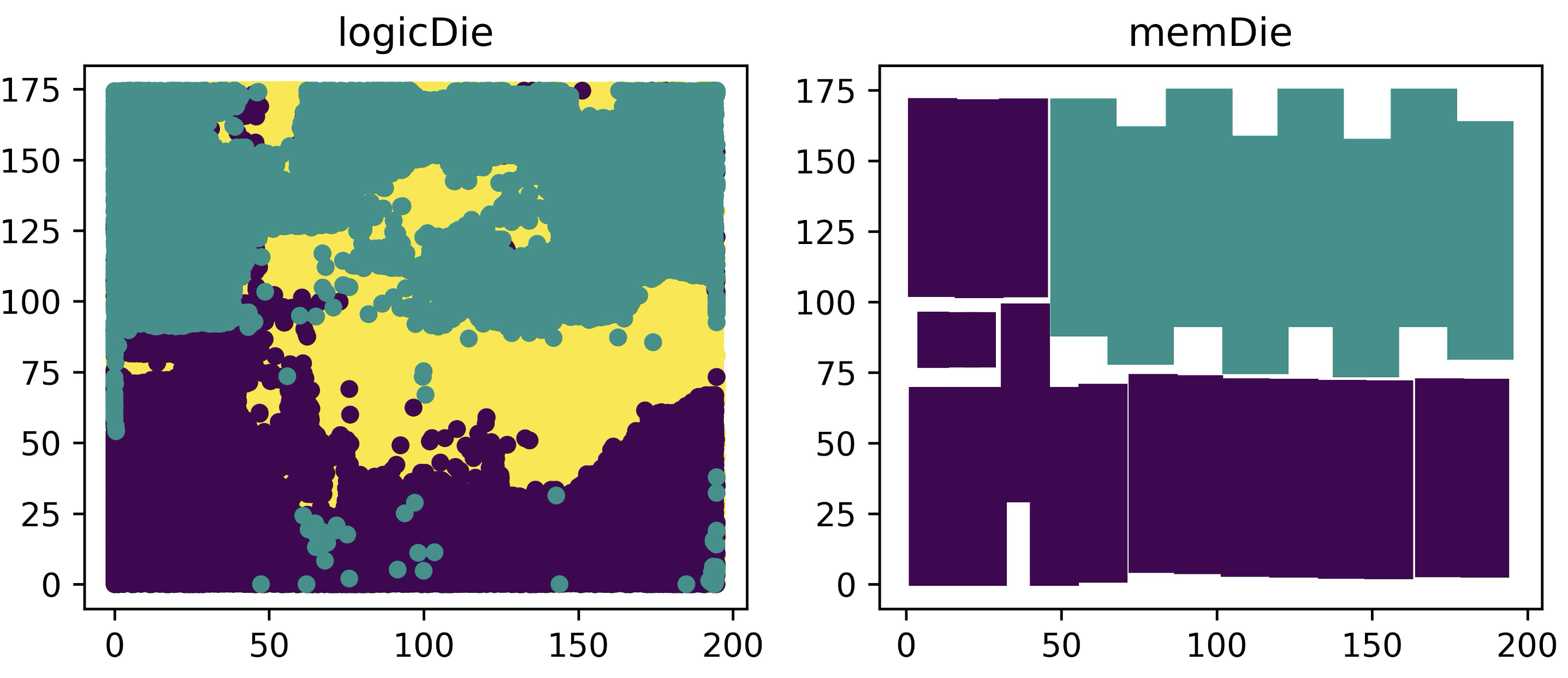}
        \caption{Base 3D placer}
    \end{subfigure}
    \begin{subfigure}[b]{0.48\textwidth}
        \centering
        \includegraphics[width=\textwidth]{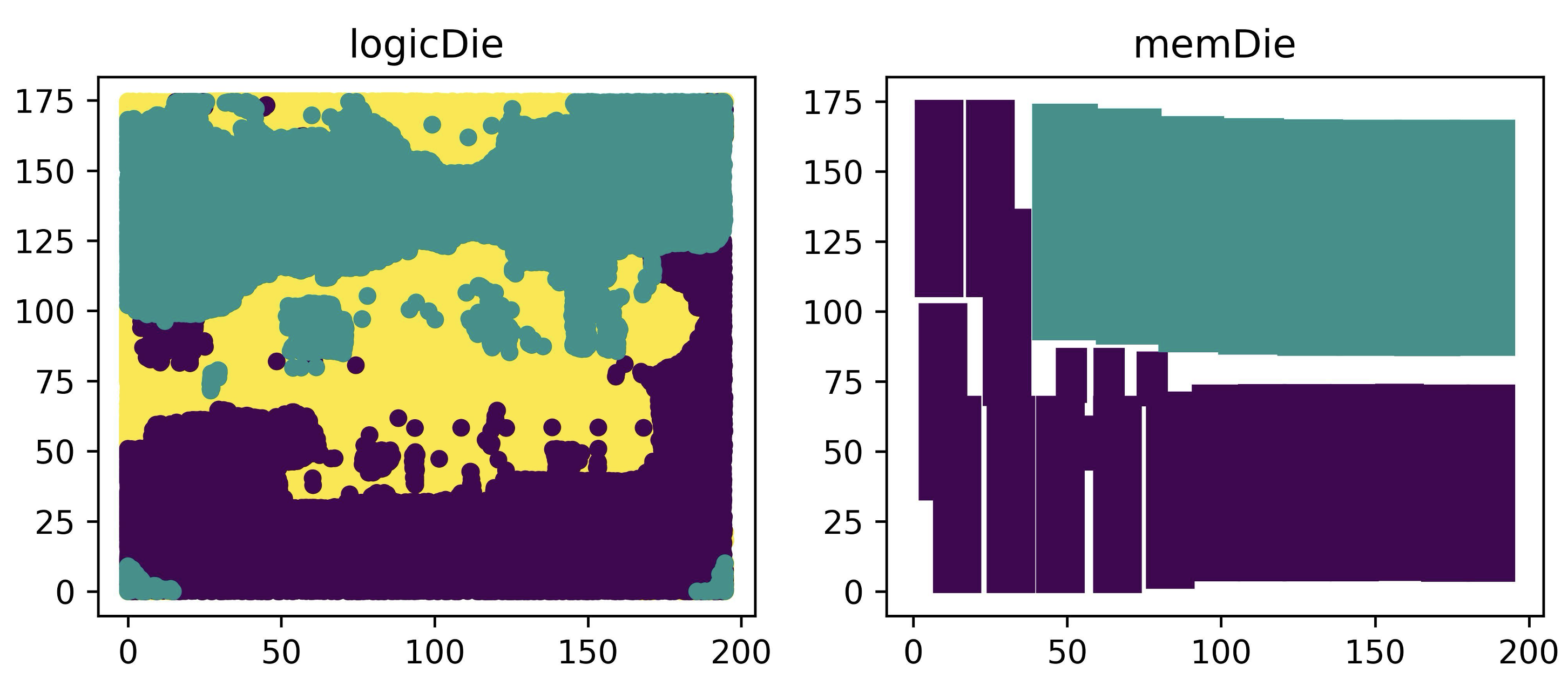} 
        \caption{EvoStage}
    \end{subfigure}
    }
    \caption{Visualization of the 3D global placement results (on both the logic die and the memory die) of the base 3D placer (left, a commercial-grade 3D chip placement tool) and EvoStage (right). Cells belonging to the same partition share the same color, and superior cohesion within clusters composed of cells of the same color indicates a higher-quality layout.}
    \label{plot:visualization}
\end{figure}

\paragraph{Results on a real-world industrial 3D chip case} Table~\ref{tab:3d} shows the desensitized GP results of the base 3D placer (i.e., the commercial-grade 3D chip placement tool) and our proposed EvoStage on the 3D chip design. Our method EvoStage significantly outperforms the original results across multiple metrics (e.g., a 9.24\% improvement of the HPWL on the logic die, and a 7.51\% improvement of the HPWL on the memory die) and achieves dramatic efficiency (i.e., a 52.21\% improvement of the optimization iterations). We also provide the desensitized visualization of the 3D global placement results in Figure~\ref{plot:visualization}, where cells belonging to the same partition share the same color (i.e., we have three partitions in this case), and superior cohesion within clusters composed of cells of the same color indicates a higher-quality layout. We can observe that EvoStage produces layouts where the standard cell clusters exhibit stronger cohesion (as shown in the logic dies in Figure~\ref{plot:visualization}, the clusters of different colors produced by EvoStage are less intermingled with each other) and the macro cells have a neater arrangement (as shown in the memory dies in Figure~\ref{plot:visualization}, the macro cells, denoted as colored rectangles, are placed by EvoStage in a neater way), which can provide more effective guidance for upstream and downstream tasks (for instance, guiding the partitioning process in floorplanning).

The great improvement in performance and efficiency achieved by EvoStage on the industrial 3D chip case indicates that it has the capability to efficiently design high-quality algorithms even for understudied tasks in real-world industry scenarios. It is worth noting that EvoStage achieves more significant improvements on the real-world industrial 3D chip case than those obtained on the open-source benchmarks in Section~\ref{sec-GP}. This is because, for open-source benchmarks, human experts have devoted many efforts and much time to iteratively refining several state-of-the-art algorithm designs, achieving highly competitive performance that leaves little room for further improvement, while in real-world industry scenarios, human experts have not conducted thorough research and do not have as much budget to iteratively refine the algorithm design through trial-and-error. This further reveals the value of EvoStage for automated algorithm design in real-world industry scenarios.



\section{Applying EvoStage to designing acquisition functions of Bayesian optimization for black-box optimization}

In order to verify the broad generalizability of our proposed method EvoStage, we extend its application to design the acquisition functions for the widely-used black-box optimization algorithm, Bayesian optimization~\cite{BOsurvey}. In Section~\ref{chip-placement}, we have applied EvoStage to design algorithm components of the famous gradient descent optimizer (i.e., the Adam optimizer~\cite{kingma2014adam}), achieving outstanding performance in the chip placement task. However, in real-world industrial scenarios, we may encounter many optimization problems where gradient information is inaccessible. These optimization problems are referred to as black-box optimization problems~\cite{bbobook}, and Bayesian optimization~\cite{BOsurvey} is an efficient and effective method for black-box optimization, with no need for gradient information. Thus, our application of EvoStage covers the two most common optimization scenarios, i.e., gradient optimization and black-box optimization.

\paragraph{Black-box optimization} Black-box optimization refers to the task of optimizing an objective function $ f: \mathcal{X} \to \mathbb{R} $ under a constrained budget for function evaluations. The term "black-box" indicates that, although we have access to $f(\boldsymbol{x})$ for any $\boldsymbol{x}\in \mathcal{X}$, no additional information about $f$ is accessible, including gradients, Hessian matrices, or other structural properties. In scenarios where function evaluations incur high costs, it becomes necessary for us to carefully and strategically select the solutions to evaluate, which has a significant impact on not only the current solution quality, but also the later optimization process. The primary goal of black-box optimization is to generate a sequence $\boldsymbol{x}_t$ that converges to the global optimum as efficiently as possible. Black-box optimization has a wide range of applications in science and engineering, such as drug discovery~\cite{terayama2021black} and material design~\cite{frazier2015bayesian}, and has attracted many efforts to develop effective methods~\cite{googlevizier,botorch}.

\paragraph{Bayesian optimization} Bayesian optimization (BO) is one of the state-of-the-art methods for black-box optimization, which is best-suited for expensive function evaluations~\cite{BOtutorial} due to its high data efficiency achieved with its ability to incorporate prior beliefs about the target problem to provide guidance for the sampling of new data, and its good balance between exploration and exploitation during the search process~\cite{BOsurvey}. The core principle of BO lies in integrating probabilistic modeling with adaptive decision-making: instead of blindly sampling evaluation points, BO leverages a probabilistic model to capture the uncertainty of the unknown objective function, then uses this uncertainty to guide the selection of the next most "informative" evaluation point--balancing the trade-off between exploration (sampling regions with high uncertainty to improve model accuracy) and exploitation (sampling regions predicted to yield optimal function values). This sequential update mechanism enables BO to efficiently converge to the global optimum with far fewer function evaluations compared to traditional black-box optimization methods (e.g., grid search, random search, and evolutionary algorithms~\cite{back1996evolutionary}). BO consists of two indispensable and iteratively coupled components, i.e., the probabilistic surrogate model and the acquisition function. 

The probabilistic surrogate model serves as a computationally cheap approximation of the expensive black-box function $f$. Its primary role is to model the mean and uncertainty of $f$ across the input space $\mathcal{X}$. The most widely used surrogate model in BO is the Gaussian process~\cite{williams1995gaussian}, due to its ability to naturally quantify uncertainty (via variance estimates) for any unevaluated input $\boldsymbol{x}\in \mathcal{X}$. Formally, a Gaussian process assumes that the function value $f(\boldsymbol{x})$ follows a multivariate normal distribution, parameterized by a mean function $m(\boldsymbol{x})$ and a covariance function $k(\boldsymbol{x}, \boldsymbol{x}')$ (e.g., squared exponential kernel). For a set of observed data $ D = \{(\boldsymbol{x}_1, f(\boldsymbol{x}_1)), ..., (\boldsymbol{x}_n, f(\boldsymbol{x}_n)) \}$,  the Gaussian process predicts the distribution of $f(\boldsymbol{x})$ at a new input $\boldsymbol{x}$ as: $ p(f(\boldsymbol{x}) \mid \boldsymbol{x}, D) = \mathcal{N}\left( \mu(\boldsymbol{x}), \sigma^2(\boldsymbol{x}) \right) $, where $\mu(\boldsymbol{x})$ represents the surrogate’s estimate of $f(\boldsymbol{x})$, and $\sigma^2(\boldsymbol{x})$ quantifies the uncertainty in this estimate--critical for guiding exploration. Other surrogate models (e.g., tree-structured Parzen estimators~\cite{watanabe2023tree} and random forests~\cite{breiman2001random}) are also used in practice, especially for high-dimensional input spaces where Gaussian processes become computationally prohibitive.

The acquisition function derives a scalar "utility score" for each unevaluated input $\boldsymbol{x}\in \mathcal{X}$; the next evaluation point $\boldsymbol{x}_{n+1}$ is selected as the input that maximizes this score. The design of the acquisition function directly embodies the exploration-exploitation trade-off, with common designs like expected improvement (EI)~\cite{EI} and upper confidence bound (UCB)~\cite{UCB}. EI computes the expected value of the improvement that $\boldsymbol{x}$ would bring over the current best observed function value, and UCB balances exploration and exploitation via a trade-off parameter $\kappa$: $ \text{UCB}(\boldsymbol{x}) = \mu(\boldsymbol{x}) + \kappa \sigma(\boldsymbol{x}) $, where a larger $\kappa$ emphasizes exploration, while a smaller $\kappa$ emphasizes exploitation.

\paragraph{Settings of EvoStage} In this experiment, we apply EvoStage to design the acquisition function, which has a great impact on the overall performance of BO as it serves as the critical "decision-making engine" that translates the probabilistic insights into actionable sampling strategies, balancing the exploration and exploitation. For the evolution process, we maintain a population of 3 algorithm individuals, reproduce 3 individuals for one generation, and make an evolution for a total of 3 generations (i.e., a total of 9 real evaluations). In this setting, an algorithm individual is just a multi-stage heuristic, corresponding to the designed acquisition function. We set the number of stages as 3 for the Stagewise-Design operator. For each designed multi-stage heuristic, we evaluate its performance by running the corresponding BO for 15 samples, i.e., each stage costs 5 samples. The implementation details can be found in Appendix~\ref{implementation-detail-GP}.

\paragraph{Open-source black-box optimization benchmarks} We conduct our experiments on open-source black-box optimization benchmarks, including synthetic problems and hyperparameter optimization problems (HPO)~\cite{eggensperger2021hpobench}. Specifically, the synthetic problems (i.e., Ackley 2D, Rastrigin 2D, Griewank 2D, and Levy 2D) come from the BoTorch synthetic functions~\cite{botorch}, which are standardized analytical benchmark problems designed for rigorous testing and comparison of BO algorithms, with diverse characteristics (e.g., multiple local optima, deceptive local optima, nearly flat outer region, steep central hole, slow convergence, etc) simulating the complex real-world problems. For the HPO benchmark, we choose the neural architecture search (NAS) problems~\cite{nas} (i.e., SliceLocalization, ProteinStructure, NavalPropulsion, and ParkinsonsTelemonitoring) to verify the outstanding performance of our method, given that these problems can better reflect real-world industrial challenges and possess great value for optimization.

\paragraph{Main results} Table~\ref{tab:bo} demonstrates the optimization results of different methods. The upper 4 benchmarks are synthetic functions where the optimal gap is calculated as the gap between the produced result and the optimal result (which is known on these synthetic functions); while the bottom 4 benchmarks are HPO-NAS problems where the optimal gap is calculated as the performance gap between the currently found architecture and the theoretically globally optimal architecture in the search space. Both metrics are the smaller the better. The metric "Pass Rate" measures the code generation quality of each LLM-based method. If a code is legal, we count it as one pass code. As shown in Table~\ref{tab:bo}, our method EvoStage designs the best acquisition function for every single case compared to the popular acquisition functions (UCB~\cite{UCB} and EI~\cite{EI}) and those designed automatically by LLM-based methods (EoH~\cite{EoH} and AlphaEvolve~\cite{alphaevolve}) with a budget of 9 evaluations (the same as EvoStage for fair comparison), even achieving an order-of-magnitude improvement on the SliceLocalization problem. Moreover, EvoStage achieves the best Pass Rate on every single case (even 100\% Pass Rate on three cases) due to the simplified code generation subtask (by the multi-stage decomposition) and the extra intermediate guidance information provided. We provide an example of the acquisition function with the best performance generated by EvoStage in Appendix~\ref{generated-algorithm} for a more intuitive understanding. Note that EoH~\cite{EoH} and AlphaEvolve~\cite{alphaevolve} achieve better performance than expert-designed acquisition functions (i.e., UCB~\cite{UCB} and EI~\cite{EI}), contrary to the observation that they struggle to generate better designs for chip placement compared to the expert-designed schedule in Section~\ref{sec-GP}. This may be because Bayesian optimization for black-box optimization has been more carefully studied than the Adam optimizer for chip placement, with LLMs encountering similar knowledge in the pre-training process.

\begin{table*}[t!]
    \caption{Optimization results obtained by popular acquisition functions (UCB~\cite{UCB} and EI~\cite{EI}) and those designed automatically by LLM-based methods (EoH~\cite{EoH}, AlphaEvolve~\cite{alphaevolve} and our EvoStage) on different benchmarks. The upper 4 benchmarks are synthetic functions, and the bottom 4 benchmarks are HPO-NAS problems. The optimal gap is the gap between the produced result and the optimal result, which is the smaller the better. The metric "Pass Rate" (i.e., rate of legal codes) measures the code generation quality of each LLM-based method. The last row shows the average rank of each method across all benchmarks with respect to the metric Optimal Gap or Pass Rate. Numbers in bold denote the best results achieved on each benchmark.}
    \label{tab:bo}
    \centering
    \resizebox{\linewidth}{!}{
    \begin{tabular}{c|c|c|cc|cc|cc}
    \toprule
    \multirow{2}{*}{Task} & \multicolumn{1}{c}{UCB~\cite{UCB}} & \multicolumn{1}{c}{EI~\cite{EI}} & \multicolumn{2}{c}{EoH~\cite{EoH}} & \multicolumn{2}{c}{AlphaEvolve~\cite{alphaevolve}} & \multicolumn{2}{c}{EvoStage (ours)}\\
    & Optimal Gap & Optimal Gap & Optimal Gap & Pass Rate(\%) & Optimal Gap & Pass Rate(\%) & Optimal Gap & Pass Rate(\%)\\
    \midrule
    Ackley 2D & 1.509e+00 & 8.459e-01 & 2.575e-01 & 66.67 & 9.325e-01 & 55.56 & \textbf{2.401e-01} & \textbf{100.00} \\
    Rastrigin 2D & 9.811e+00 & 7.441e+00 & 2.733e+00 & 66.67 & 2.311e+00 & 11.11 & \textbf{1.927e+00} & \textbf{100.00} \\
    Griewank 2D& 1.370e+00 & 6.617e-01 & 4.349e-01 & 77.78 & 1.337e-01 & 55.56 & \textbf{1.211e-01} & \textbf{88.89} \\
    Levy 2D & 3.035e+00 & 7.125e-03 & 3.755e-02 & \textbf{44.44} & 7.200e-03 & 22.22 & \textbf{1.228e-03} & \textbf{44.44} \\ \midrule
    SliceLocalization & 1.527e-03 & 4.154e-03 & 1.089e-03 & 55.56 & 2.000e-03 & 44.44 & \textbf{2.389e-04} & \textbf{88.89} \\
    ProteinStructure & 3.492e-01 & 3.492e-01 & 2.605e-01 & 66.67 & 2.463e-01 & 66.67 & \textbf{2.324e-01} & \textbf{88.89} \\
    NavalPropulsion & 6.146e-04 & 4.779e-03 & 5.624e-05 & 77.78 & 1.410e-02 & 33.33 & \textbf{3.191e-05} & \textbf{88.89} \\
    ParkinsonsTelemonitoring & 3.219e-02 & 4.757e-02 & 1.108e-02 & 66.67 & 2.400e-02 & 77.78 & \textbf{9.426e-03} & \textbf{100.00} \\
    \midrule
    Avg Rank & 4.25 & 3.875 & 2.625 & 2 & 3.125 & 2.625 & \textbf{1} & \textbf{1}\\
    \bottomrule
    \end{tabular}
    }
\end{table*}

In fact, LLM-based algorithm design methods have been utilized to enhance BO. FunBO~\cite{funbo} uses the LLM to generate acquisition functions based on FunSearch~\cite{FunSearch}, and EvolCAF~\cite{evolcaf} adopts the same idea and extends it to the cost-aware acquisition function design. In this work, we have compared our proposed EvoStage with the enhanced version of FunSearch, i.e., AlphaEvolve~\cite{alphaevolve}, showing that EvoStage is significantly better. It is also worth noting that the previous works~\cite{funbo,evolcaf} focus on deriving a new acquisition function on an offline problem instance set to get an average best result on the target problem instance set, but in real-world industry scenarios, we often hope to achieve the best optimization results for every target case instead of the average best results. As shown empirically, there is no one specific acquisition function that can offer optimal solutions with search efficiency, on every function landscape~\cite{hoffman2011portfolio}, similar to the well-known No Free Lunch Theorem~\cite{wolpert2002no}. Furthermore, in real-world industry scenarios, we are usually not provided with an offline problem instance set with sufficient problem instances similar to our target problem instances. Therefore, our setting focuses on an online acquisition function design, where we utilize the LLMs to adaptively generate acquisition functions according to the target problem instance at hand, which can achieve better performance in practice.

\begin{figure}[t]
\centering
    \includegraphics[width=\textwidth]{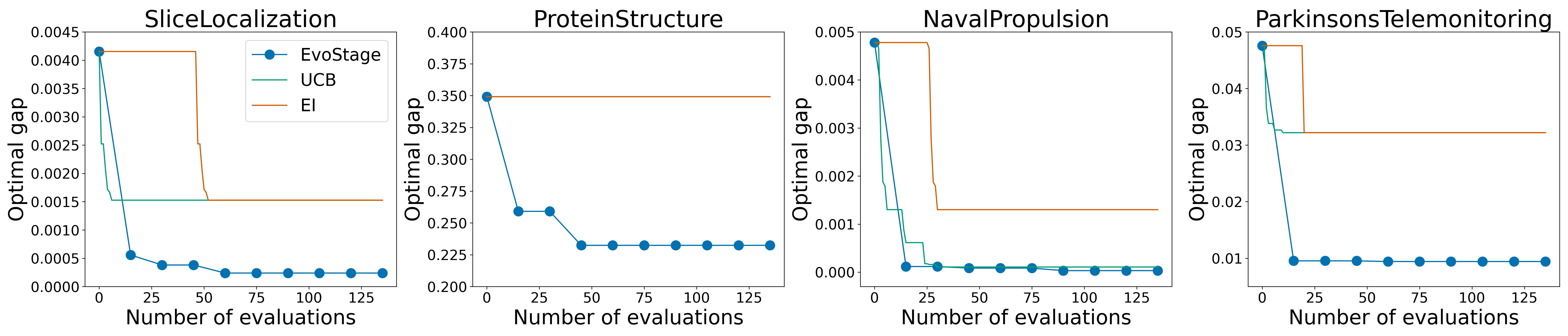} 
\caption{The optimal gap curves of different acquisition functions (UCB~\cite{UCB} and EI~\cite{EI}) and the online designing of our method EvoStage on 4 NAS problems. In the ProteinStructure problem, the curve of UCB coincides with that of EI. The points on the curve of EvoStage indicate the current best optimal gap achieved by the best acquisition design so far.}
\label{fig:bo}
\end{figure}

\paragraph{Comparison with popular acquisition functions under the same sampling cost} 
Despite exhibiting superior performance to directly employing the popular acquisition functions (i.e., UCB~\cite{UCB} and EI~\cite{EI}), EvoStage actually incurs a higher sampling cost: we adopt 9 rounds of evolution (i.e., reproduce 9 individuals in the evolution process) to design the acquisition function, where each acquisition function requires 15 sampling operations for evaluation. That is, the total sampling cost amounts to 9$\times$15. Note that directly using a popular acquisition function incurs a sampling cost of 15. Thus, a natural and interesting question arises as to whether EvoStage can still maintain its superiority if the sampling budget of directly using a popular acquisition function reaches 9$\times$15. Figure~\ref{fig:bo} demonstrates the optimal gap curves of using different acquisition functions and the online designing of our method EvoStage on 4 NAS problems. The points on the curve of EvoStage indicate the current best optimal gap achieved by the best acquisition design, since every acquisition function is evaluated from scratch, and we do not plot the detailed curve of each acquisition function for a clear comparison. Note that in each evolution step, the new acquisition function generated by EvoStage is evaluated from scratch, and no previous evaluation points from the previous evolution process are provided to the LLM agents. Therefore, the outstanding performance achieved by EvoStage does not result from additional data point sampling, but rather stems from the practical experience summarized from the algorithm’s runtime feedback information, which enables the design of a more effective acquisition function tailored to the current problem. As shown in Figure~\ref{fig:bo}, EvoStage achieves the best results on all cases, with significant efficiency with respect to the same budget of evaluations.

\section{Conclusion}

In this work, we seek to advance automated algorithm design with LLMs, enabling it for increasingly complex tasks in industry-level scenarios. To achieve that, we propose Evolutionary Stagewise Algorithm Design (EvoStage), a new paradigm that better aligns with the capabilities of LLMs and boosts LLMs’ performance in algorithm design tasks, especially in real-world industry-level scenarios with limited budgets, by automatically decomposing the complex design task into simpler subtasks of multiple stages, solving the subtasks stage-by-stage, and evolving such solutions for a more novel algorithm with higher quality. Under this paradigm, we further introduce a multi-agent system, enabling the system to reduce the algorithm design space by coordinating different LLM agents to design different algorithm components, and a "global-local perspective" mechanism, balancing the local stage optimization and the global overall optimization to avoid falling into local optima. We apply EvoStage to the design of two types of widely used optimization methods (i.e., gradient-based optimization and black-box optimization): Adam, a famous gradient-based optimizer, and Bayesian optimization, a famous black-box optimization method. Empirical results from open-source benchmarks to a real-world industrial 3D chip design demonstrate the outstanding performance and efficiency of EvoStage, beating expert-designed algorithms within just a couple of iterations, as well as other LLM-based methods, and achieving dramatic improvements on the pioneering chip design.

We hope the proposed paradigm EvoStage can stand as a milestone along the path of automated algorithm design with LLMs, and can be extensively adopted in industry in the future, ultimately substantially elevating human productivity. Despite its exceptional capability in designing iterative algorithms that are widely applied in real-world industrial scenarios, the proposed Stagewise Design degenerates into a traditional black-box modeling algorithm design when confronted with scenarios involving one-time execution algorithms or those where the algorithm cannot be paused for information acquisition during execution, which calls for a more efficient information exploitation in these scenarios for future works. Moreover, as the capability of the base LLMs grows rapidly in the future, the intermediate information could contain information from more modalities (e.g., the 3D visualizations of a 3D chip placement), also requiring more efficient communication and collaboration among multiple LLM agents with different expertise.

\section*{Acknowledgments}
This work was supported by the National Science and Technology Major Project (2022ZD0116600), the Jiangsu Science Foundation Leading-edge Technology Program (BK20232003), the Fundamental Research Funds for the Central Universities (14380020), and the National Science Foundation of China (624B2069). The authors want to acknowledge support from the Huawei Technology Cooperation Project.

\newpage
\bibliographystyle{unsrt}  
\bibliography{references}  

\newpage

\appendix

\section{Implementation details of EvoStage} 

\subsection{Designing parameter configuration schedules of Adam optimizer for chip placement} \label{implementation-detail-GP}

\begin{figure}[h]
\centering
\includegraphics[width=\textwidth]{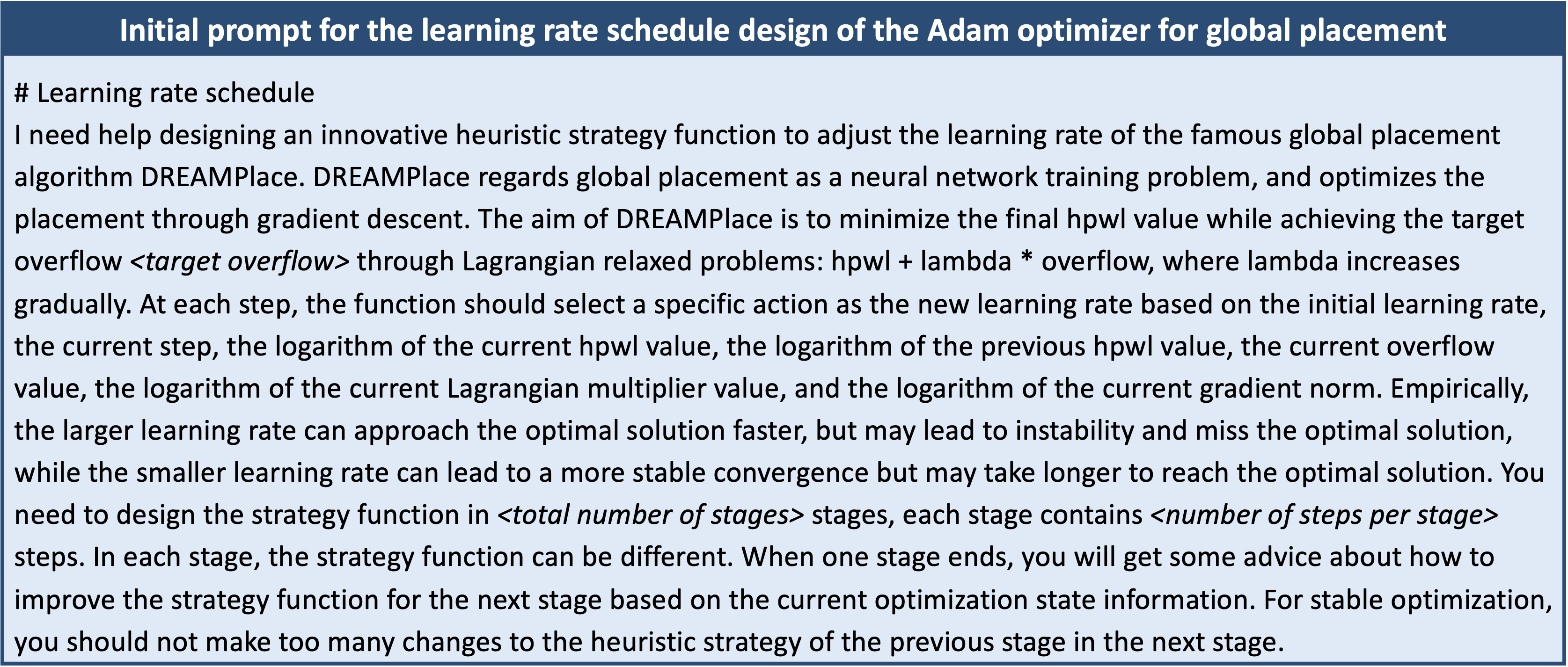}
\caption{The initial prompt for the learning rate schedule design of the Adam optimizer for global placement.}
\label{plot:prompt-lr}
\end{figure}

\begin{figure}[h]
\centering
\includegraphics[width=\textwidth]{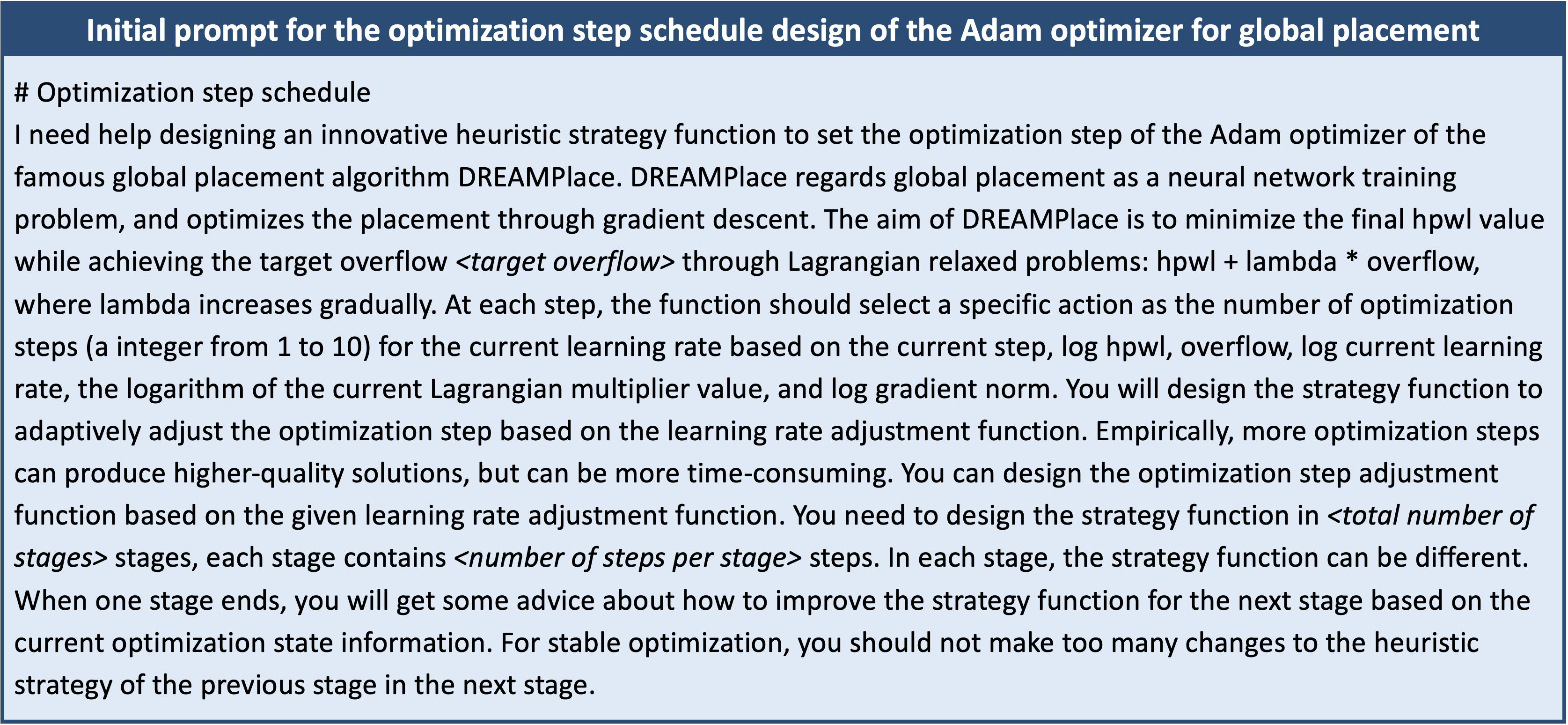}
\caption{The initial prompt for the optimization step schedule design of the Adam optimizer for global placement.}
\label{plot:prompt-os}
\end{figure}

\begin{figure}[h]
\centering
\includegraphics[width=\textwidth]{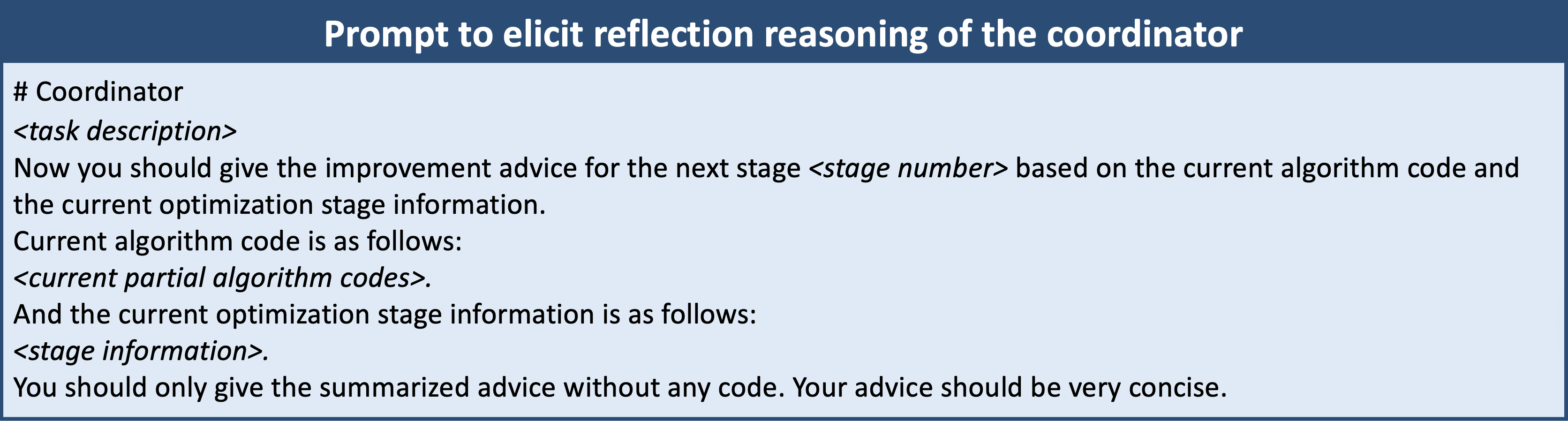}
\caption{Prompt to elicit the reflection reasoning of the coordinator.}
\label{plot:prompt-coordinator}
\end{figure}

Figures~\ref{plot:prompt-lr} and~\ref{plot:prompt-os} demonstrate the detailed initial prompts (i.e., the prompts for stage 0 of the Stagewise-Design operator) used by EvoStage for the learning rate schedule design and the optimization step schedule design, respectively. In the initial prompts, we provide the LLMs with a description of the global placement task with a detailed background of how global placement is optimized, and some domain knowledge summarized by human experts (e.g., how learning rate impacts the optimization process of global placement) to boost their reasoning ability. Prompts to elicit the stagewise algorithm design (i.e., the subsequent stages) can be found in Figure~\ref{plot:prompt-example}, and the prompt to elicit reflection reasoning of the coordinator is shown in Figure~\ref{plot:prompt-coordinator}. Prompts of the two global perspective operators can be found in Figure~\ref{plot:global}.


Table~\ref{tab:parameter-GP} gives the hyperparameter setting of EvoStage for the global placement task. We use gpt-4o as the LLM model for both the LLM coder agents and the LLM coordinator agent. We set the temperature of the LLM coder agents as 0.2 for a stable generation of valid codes and the temperature of the LLM coordinator agents as 0.7 to encourage creative reflection generations. As mentioned in Section~\ref{sec-GP}, for the evolution process, we maintain a population of 5 algorithm individuals, reproduce 5 individuals for one generation, and make an evolution for a total of 5 generations (i.e., a total of 25 real evaluations). We set the selection number $k$ for the global perspective operator Global-Explore as 2 (i.e., the number of references selected for applying the Global-Explore operator to generate a new algorithm individual), and the number of stages for the Stagewise-Design operator as 4. We initialize the population with multi-stage heuristics. For the natural language descriptions proposed in EoH~\cite{EoH}, which are evolved together with the codes of heuristics to provide the LLMs with more understanding of the heuristics, we turn them off in the code generation procedure of LLM coder agents (i.e., set the thought-of-code as False). The reason is that the natural language descriptions in the code generation procedure of LLM coder agents will exacerbate the hallucination since LLMs have not encountered similar scenarios in the pre-training process and give descriptions inconsistent with the actual performance of the codes. Supporting evidence for this can be found in the experimental results in Section~\ref{sec-GP}, where EoH~\cite{EoH} achieves worse results than algorithms designed by human experts due to the exacerbated hallucination caused by the natural language descriptions.


\begin{table*}[h]
    \caption{The hyperparameter settings of EvoStage for designing parameter configuration schedules of the Adam optimizer for chip placement.}
    \label{tab:parameter-GP}
    \centering
    \begin{tabular}{lclc}
    \toprule
     Parameter & Value & Parameter & Value \\
     \midrule
     LLM settings\\
     \midrule
     LLM coder model & gpt-4o & LLM coordinator model & gpt-4o\\
     LLM coder temperature & 0.2 & LLM coordinator temperature & 0.7\\
     \midrule
     Evolution settings\\
     \midrule
     Number of generations & 5 & Population size & 5 \\
     Selection number & 2 & Number of stages & 4 \\
     \midrule
     EvoStage settings\\
     \midrule
     Multi-stage initialization & True & Thoughts of code & False\\
    \bottomrule
    \end{tabular}
\end{table*}

\subsection{Designing acquisition functions of Bayesian optimization for black-box optimization} \label{implementation-detail-BO}
\begin{figure}[h]
\centering
\includegraphics[width=\textwidth]{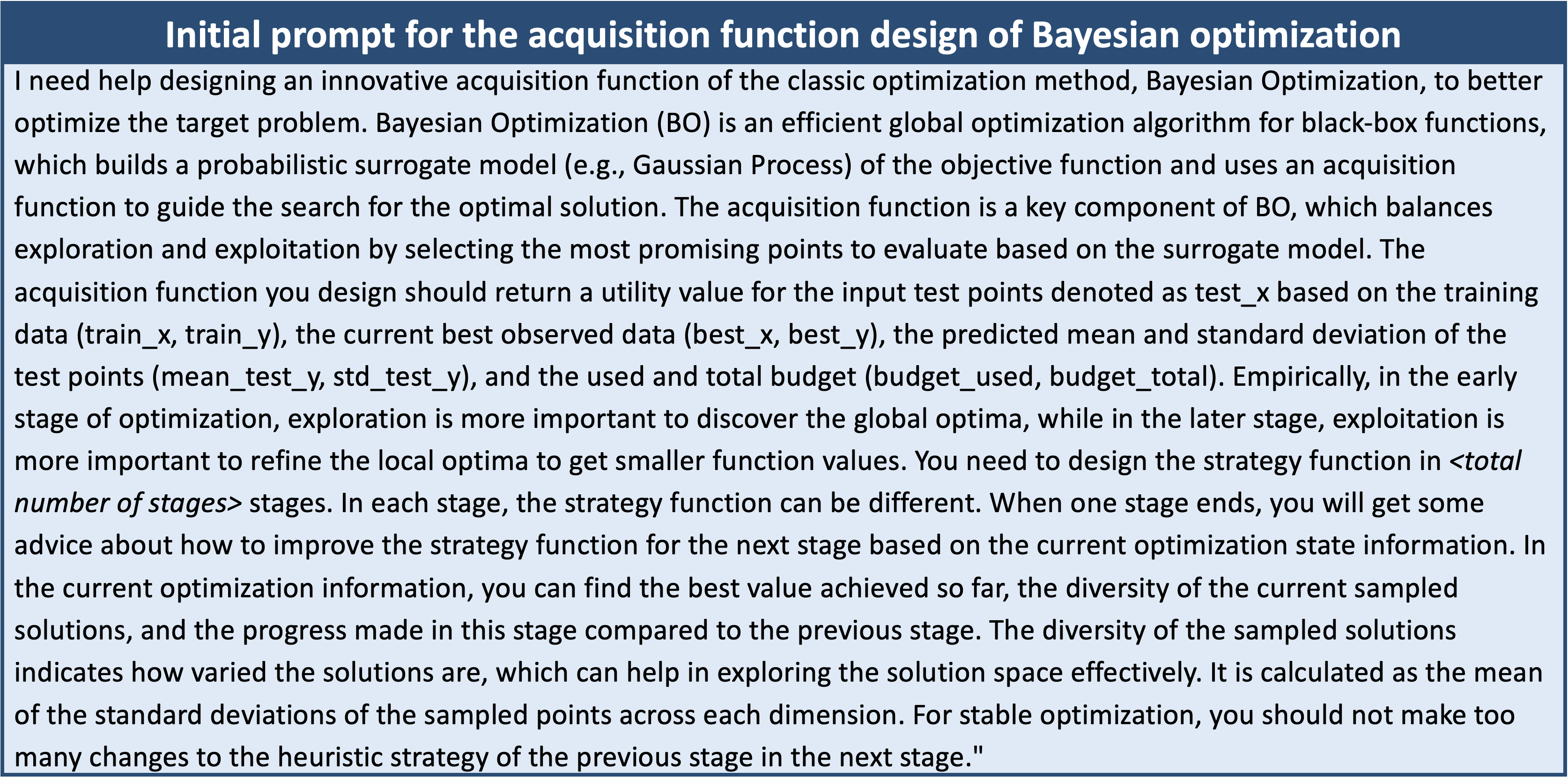}
\caption{The initial prompt for the acquisition function design of Bayesian optimization.}
\label{plot:prompt-af}
\end{figure}

For the task of Bayesian optimization, we provide the prompts for the acquisition function design task of Bayesian optimization and the detailed hyperparameter settings of EvoStage. Figure~\ref{plot:prompt-af} demonstrates the detailed initial prompts for the acquisition function design. In the initial prompts, we provide the LLMs with a description of the acquisition function design task with a detailed background of Bayesian optimization and acquisition function, and some domain knowledge summarized by human experts (e.g., how human experts typically design acquisition functions) to boost their reasoning ability. Since the prompts to elicit the stagewise algorithm design, the prompt to elicit reflection reasoning of the coordinator, and the prompts of the two global perspective operators used in the GP task can be adapted to diverse tasks merely by adjusting task-related information, we extend them for the task of Bayesian optimization, which have been shown in Figure~\ref{plot:prompt-example}, Figure~\ref{plot:prompt-coordinator}, and Figure~\ref{plot:global}, respectively. Table~\ref{tab:parameter-BO} gives the hyperparameter setting for the Bayesian optimization task. We apply the same LLM settings as the GP task, and for the evolution process, we maintain a population of 3 algorithm individuals, reproduce 3 individuals for one generation, and make an evolution for a total of 3 generations (i.e., a total of 9 real evaluations). We set the selection number $k$ for the global perspective operator Global-Explore as 2 (i.e., the number of references selected for applying the Global-Explore operator to generate a new algorithm individual), and the number of stages for the Stagewise-Design operator as 3. We initialize the population with multi-stage heuristics. Note that, contrary to the settings in the GP task, we set the thought-of-code as True to apply the natural language descriptions and find better performance on the Bayesian optimization task. This is because the acquisition function is more thoroughly studied than the GP task, and the LLM coder agents may have a certain understanding of the acquisition function generation task from the pre-training process, without the guidance from the LLM coordinator agent. Therefore, applying the natural language description can help the LLM coder agents to gain a better overall understanding of the code generation task, as well as having more ideas to generate high-quality code.

\begin{table*}[h]
    \caption{The hyperparameter settings of EvoStage for designing acquisition functions of Bayesian optimization.}
    \label{tab:parameter-BO}
    \centering
    \begin{tabular}{lclc}
    \toprule
     Parameter & Value & Parameter & Value \\
     \midrule
     LLM settings\\
     \midrule
     LLM coder model & gpt-4o & LLM coordinator model & gpt-4o\\
     LLM coder temperature & 0.2 & LLM coordinator temperature & 0.7\\
     \midrule
     Evolution settings\\
     \midrule
     Number of generations & 3 & Population size & 3 \\
     Selection number & 2 & Number of stages & 3 \\
     \midrule
     EvoStage settings\\
     \midrule
     Multi-stage initialization & True & Thoughts of code & True\\
    \bottomrule
    \end{tabular}
\end{table*}

\section{Detailed codes of algorithms designed by EvoStage}\label{generated-algorithm}
\begin{figure}[h]
\centering
    \includegraphics[width=\textwidth]{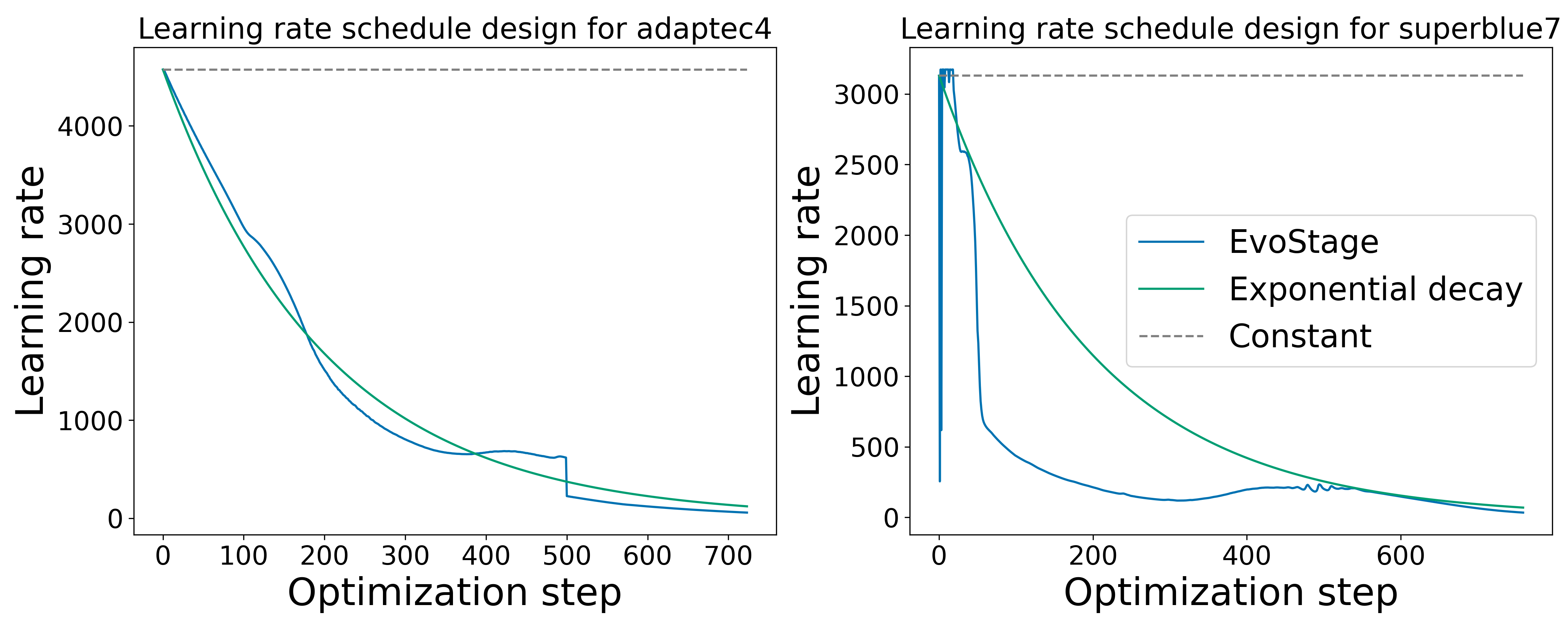} 
\caption{The curve of the learning rate schedules designed by EvoStage, compared to those of the two most widely-used schedules (i.e., exponential decay and constant schedules) on the two chip cases, adaptec4 and superblue7.}
\label{plot:lr-curve}
\end{figure}

\begin{figure}[h]
\centering
    \begin{subfigure}[b]{\textwidth}
    \centering
    \includegraphics[width=\textwidth]{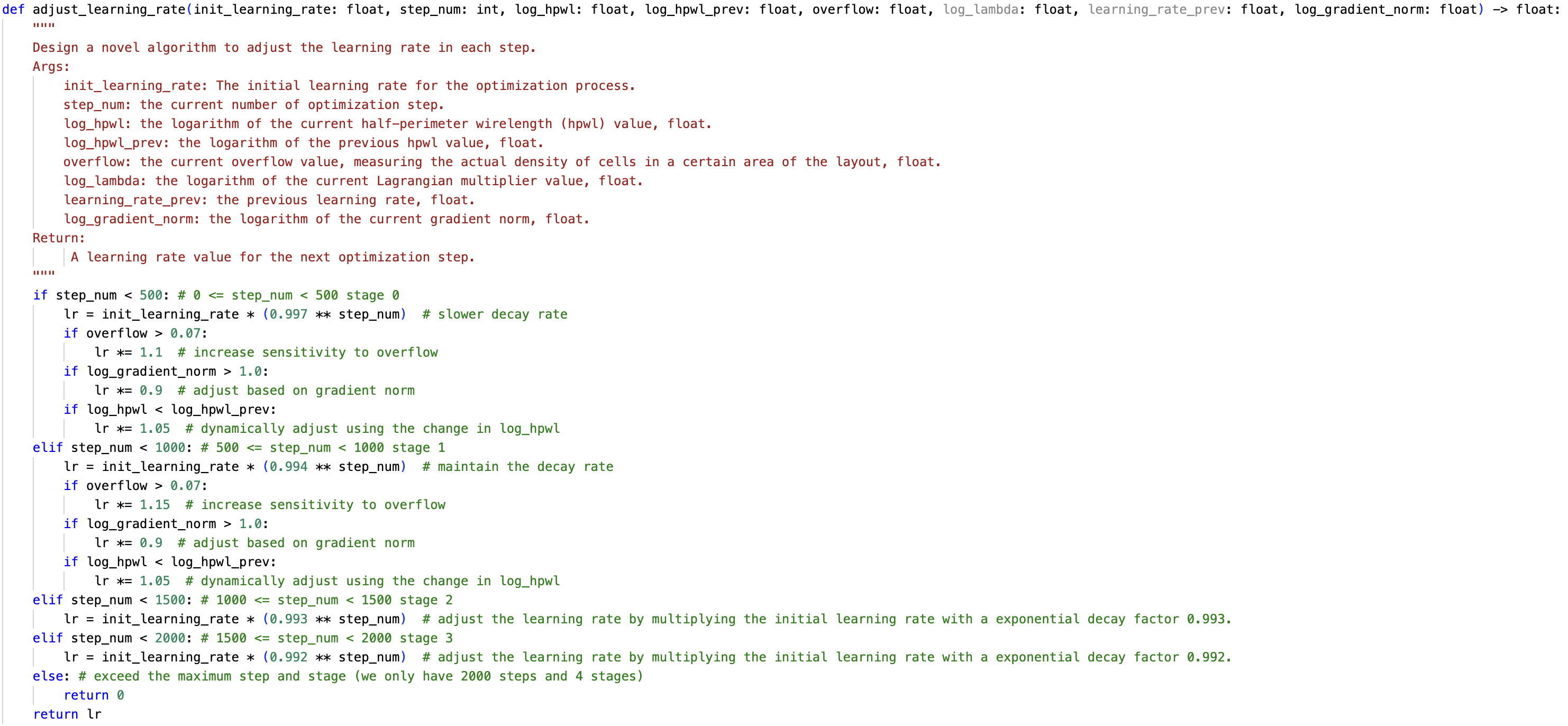} 
    \subcaption{Generated learning rate schedule 1 for the Adam optimizer in the GP task}
    \label{lr1}
    \end{subfigure}
    \begin{subfigure}[b]{\textwidth}
    \centering
    \includegraphics[width=\textwidth]{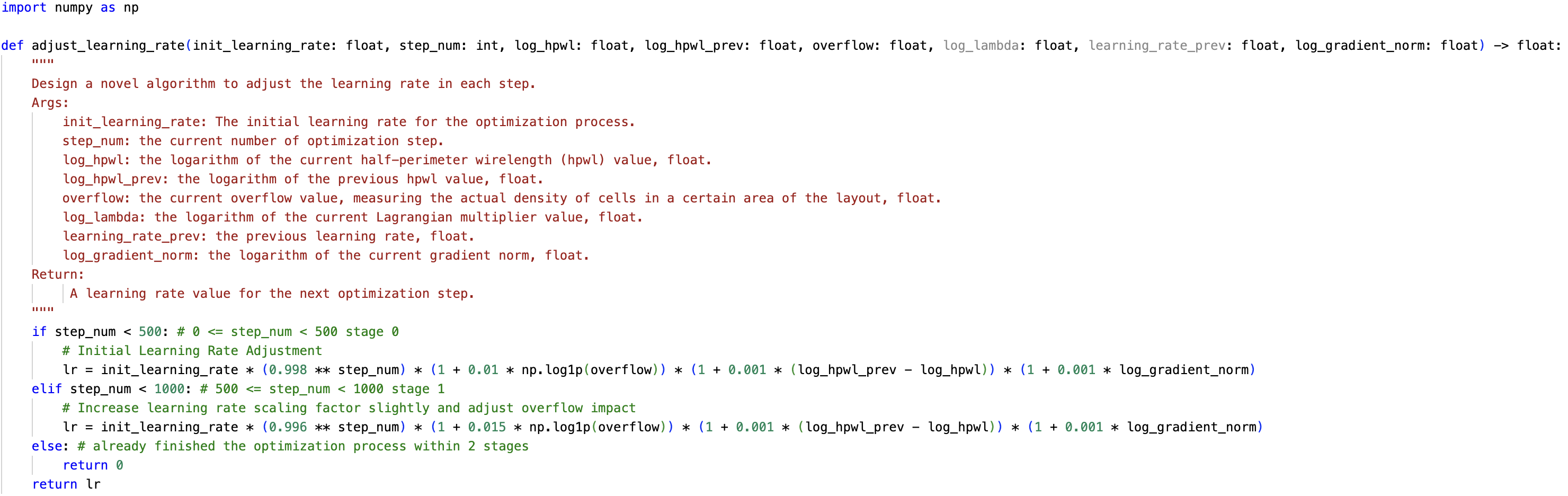} 
    \subcaption{Generated learning rate schedule 2 for the Adam optimizer in the GP task}
    \label{lr2}
    \end{subfigure}
    \begin{subfigure}[b]{0.48\textwidth}
    \centering
    \includegraphics[width=\textwidth]{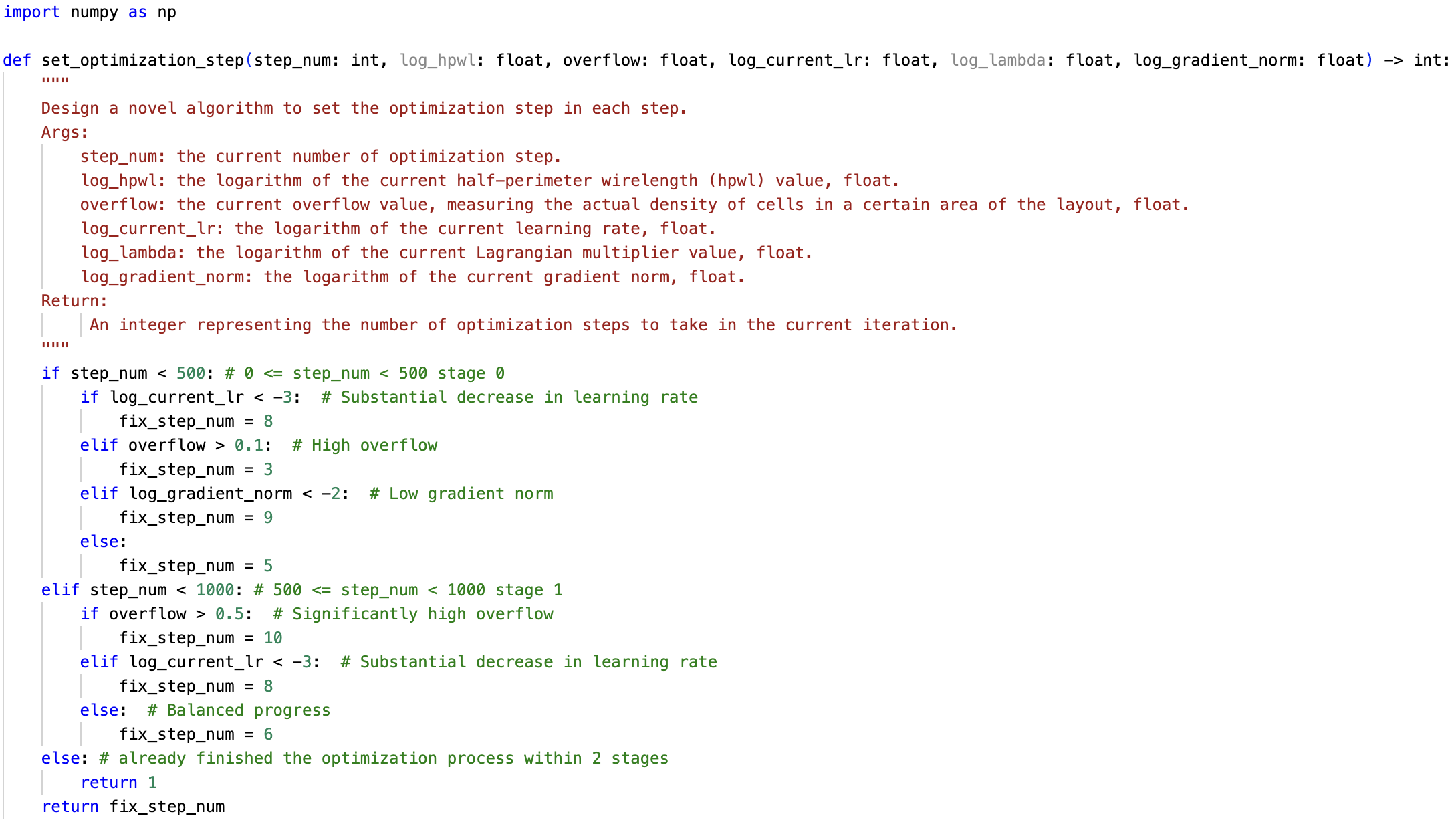}
    \subcaption{Generated optimization step schedule for the Adam optimizer in the GP task}
    \label{os1}
    \end{subfigure}
    \begin{subfigure}[b]{0.48\textwidth}
    \centering
    \includegraphics[width=\textwidth]{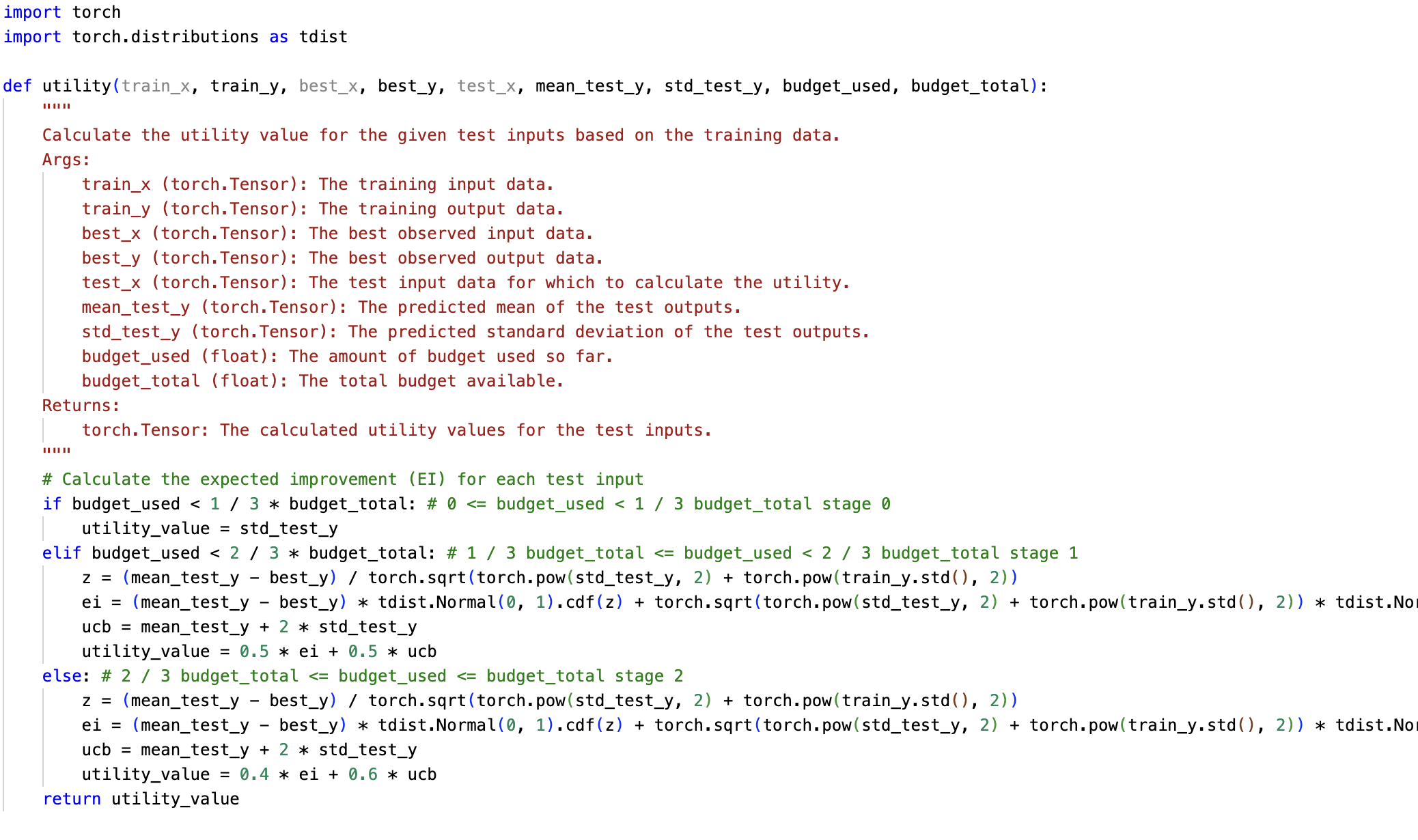}
    \subcaption{Generated acquisition function for Bayesian optimization}
    \label{af}
    \end{subfigure}
\caption{Examples of the algorithm codes generated by our proposed EvoStage.}
\label{plot:code-demo}
\end{figure}

In this section, we provide examples of the algorithm codes generated by our method EvoStage. As shown in Figure~\ref{lr1} to~\ref{os1}, for the GP task, the generated learning rate schedules and optimization step schedules contain different stages (e.g., if step\_num < 500: \#stage 0 $\dots$ elif step\_num < 1000 \#stage 1 $\dots$). They can make full use of the current intermediate optimization information (e.g., HPWL, overflow, gradient norm, etc). For example, the learning rate schedule function is defined as 
def adjust\_learning\_rate(init\_learning\_rate: float, step\_num: int, log\_hpwl: float, log\_hpwl\_prev: float, overflow: float, log\_lambda: float, learning\_rate\_prev: float, log\_gradient\_norm: float) -> float, where all of the parameters are used in the adjustment except for log\_lambda and learning\_rate\_prev. We can observe that the generated schedules are able to adaptively adjust the corresponding parameters according to the current optimization information (e.g., if overflow > 0.7, log\_gradient\_norm > 1.0, and if log\_hpwl < log\_hpwl\_prev, etc), just like an experienced expert. We plot the curve of the designed learning rate schedules for the two chip cases (i.e., adaptec4 and superblue7) in Figure~\ref{plot:lr-curve}. Compared to the most widely-used exponential decay schedules (with decay factor as 0.995) and the constant schedules, the learning rate schedules designed by EvoStage vary across different cases and demonstrate the capacity of adaptive adjustment under different optimization situations (e.g., increase the learning rate or keep it stable when necessary). For the task of designing acquisition functions of Bayesian optimization for black-box optimization, we can observe from Figure~\ref{af} that the generated acquisition function contains an interesting and reasonable idea: it encourages exploration in the early stage by directly using the predicted standard deviation of the test outputs as the utility value (i.e., utility\_value = std\_test\_y), and places more emphasis on exploitation by using a weighted utility value consisting of EI and UCB (i.e., utility\_value = 0.4 * ei + 0.6 * UCB).

\section{Visualization of global placement layouts}~\label{layout-plots}

\begin{figure}[h]
\centering
    \begin{subfigure}[b]{0.48\textwidth}
    \centering
    \includegraphics[width=\textwidth]{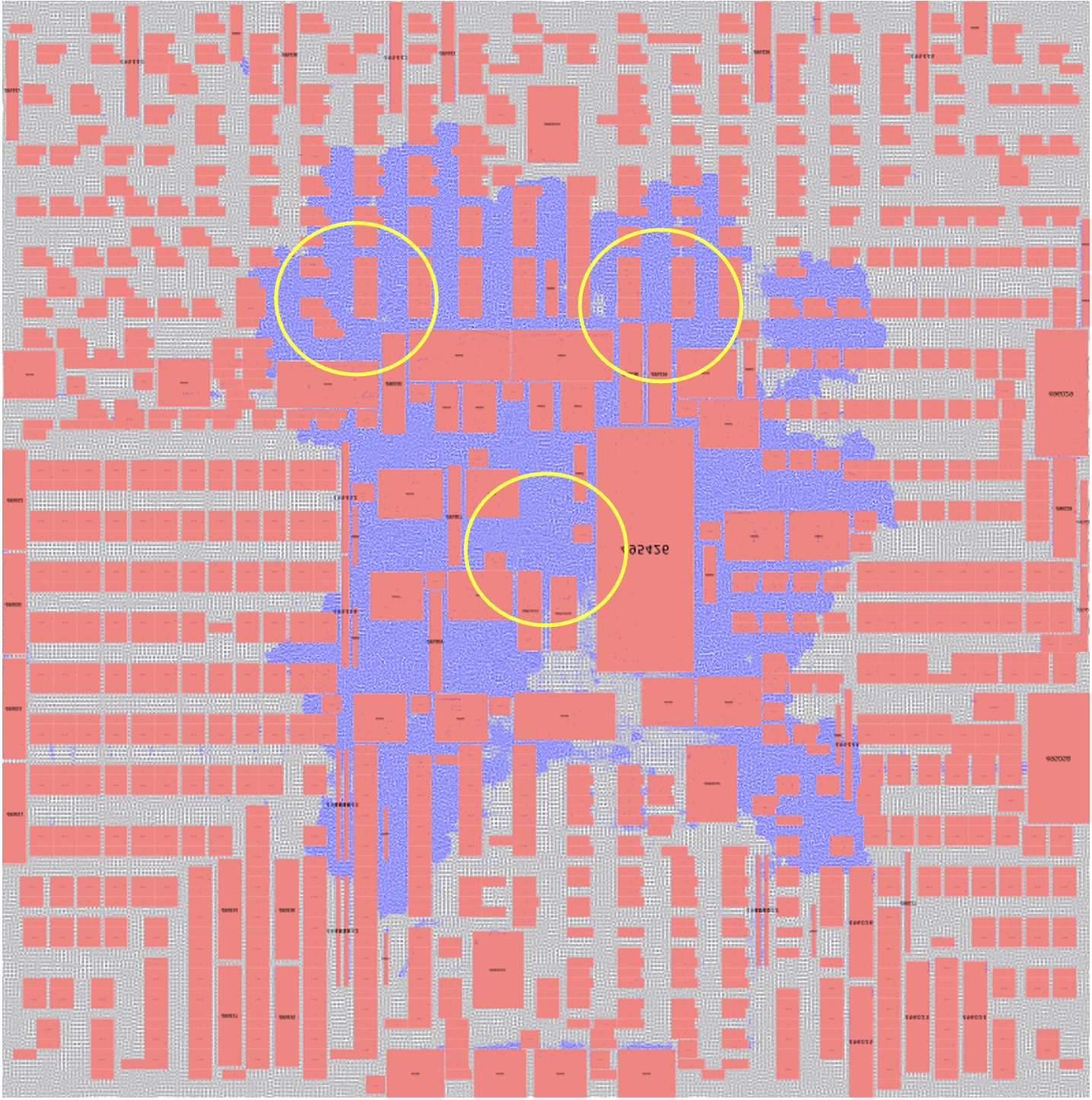} 
    \subcaption{GP plot of EvoStage}
    \end{subfigure}
    \begin{subfigure}[b]{0.48\textwidth}
    \centering
    \includegraphics[width=\textwidth]{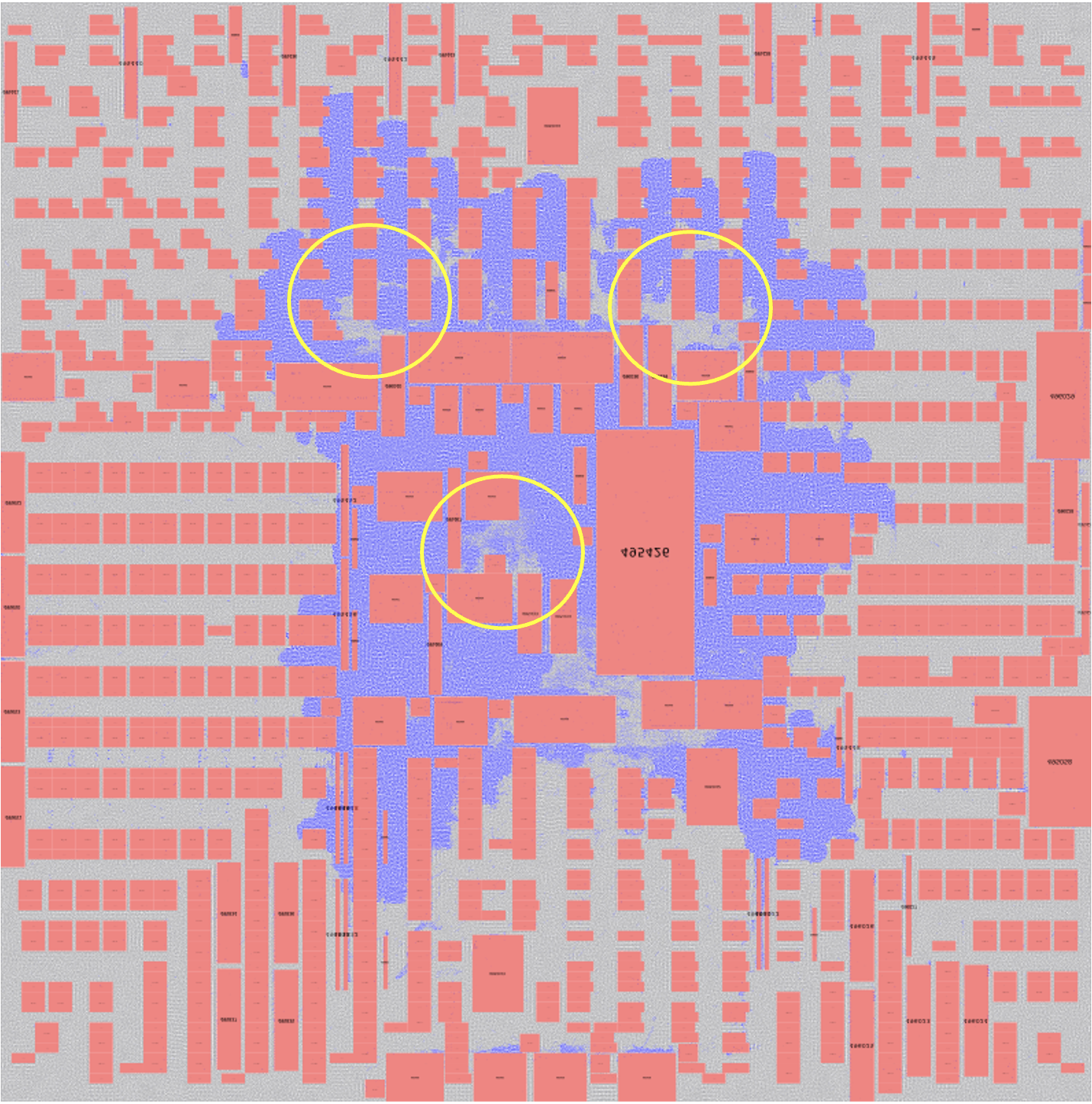}
    \subcaption{GP plot of DREAMPlace-Nesterov}
    \end{subfigure}
\caption{The global placement plot comparison of EvoStage and DREAMPlace-Nesterov~\cite{lin2020dreamplace} on adaptec4, where movable standard cells are denoted in blue and fixed macro cells are denoted in red. The critical areas, where EvoStage produces better layouts (e.g., neater boundaries or stronger cohesion), are highlighted in yellow circles.}
\label{plot:adaptec4}
\end{figure}

\begin{figure}[h]
\centering
    \begin{subfigure}[b]{0.48\textwidth}
    \centering
    \includegraphics[width=\textwidth]{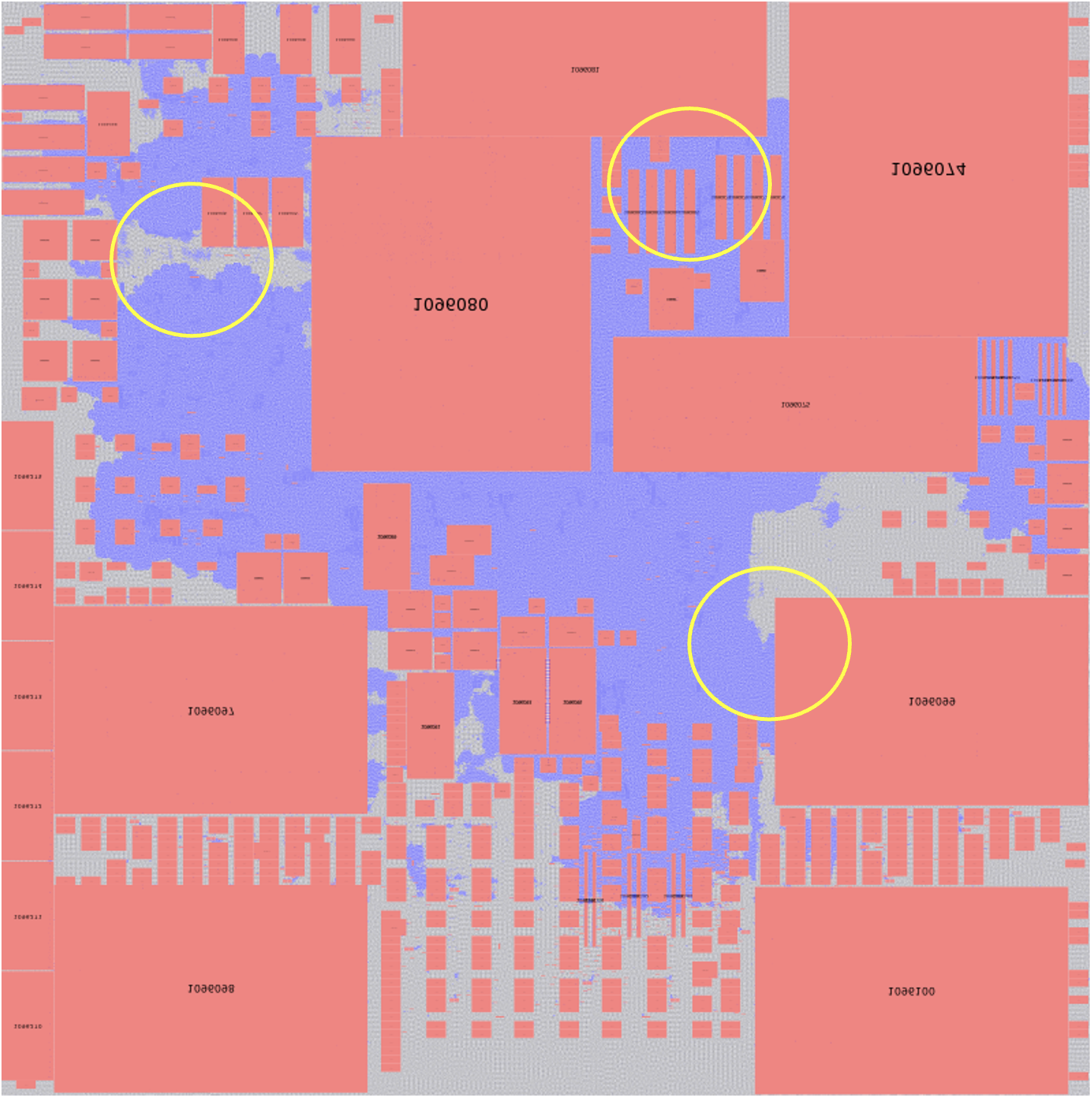} 
    \subcaption{GP plot of EvoStage}
    \end{subfigure}
    \begin{subfigure}[b]{0.48\textwidth}
    \centering
    \includegraphics[width=\textwidth]{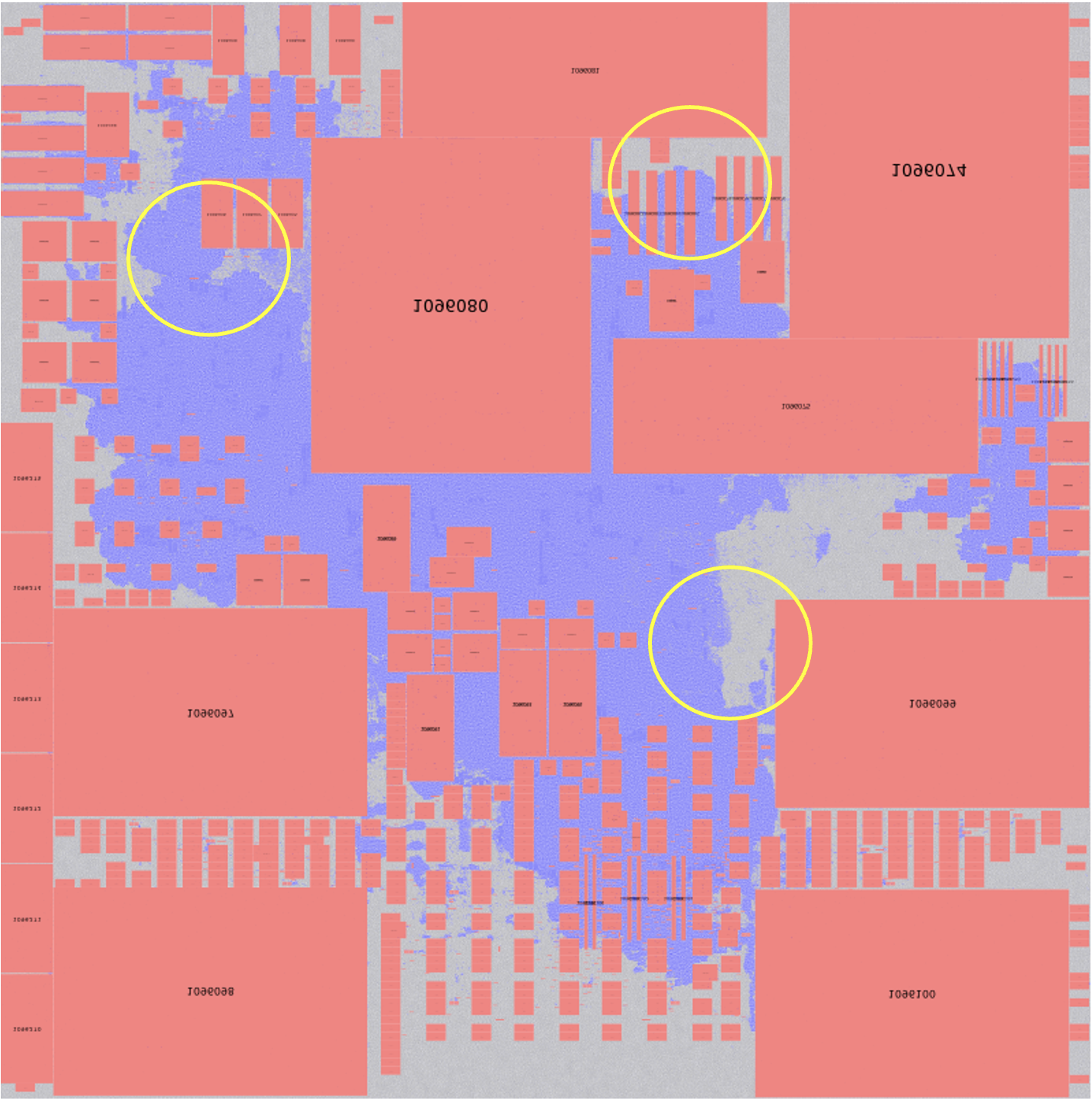}
    \subcaption{GP plot of DREAMPlace-Nesterov}
    \end{subfigure}
\caption{The global placement plot comparison of EvoStage and DREAMPlace-Nesterov~\cite{lin2020dreamplace} on bigblue3, where movable standard cells are denoted in blue and fixed macro cells are denoted in red. The critical areas, where EvoStage produces better layouts (e.g., neater boundaries or stronger cohesion), are highlighted in yellow circles.}
\label{plot:bigblue3}
\end{figure}

\begin{figure}[h]
\centering
    \begin{subfigure}[b]{0.48\textwidth}
    \centering
    \includegraphics[width=\textwidth]{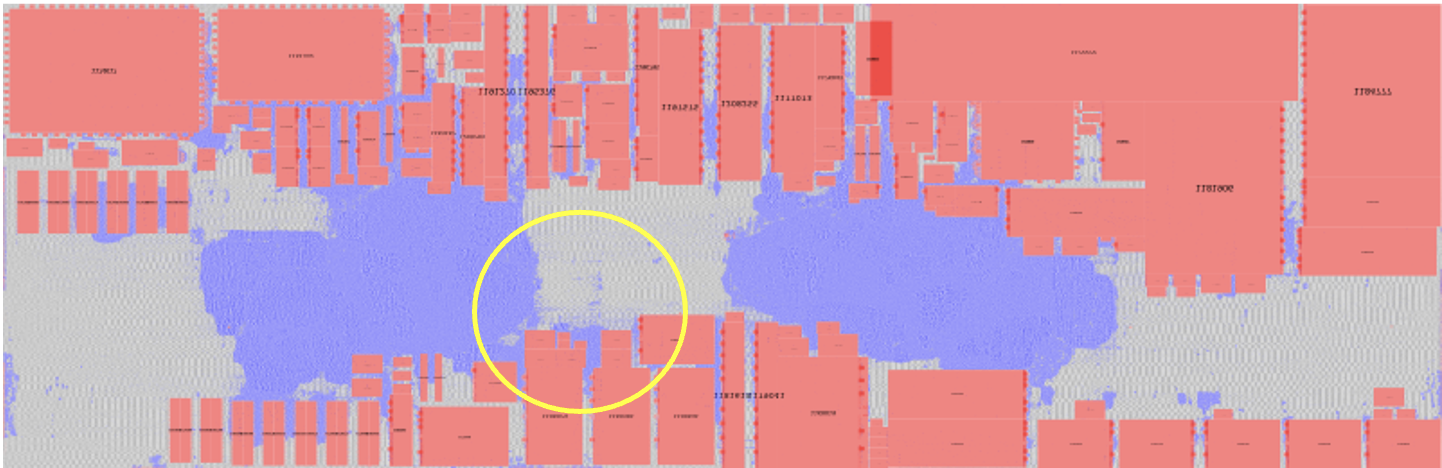} 
    \subcaption{GP plot of EvoStage}
    \end{subfigure}
    \begin{subfigure}[b]{0.48\textwidth}
    \centering
    \includegraphics[width=\textwidth]{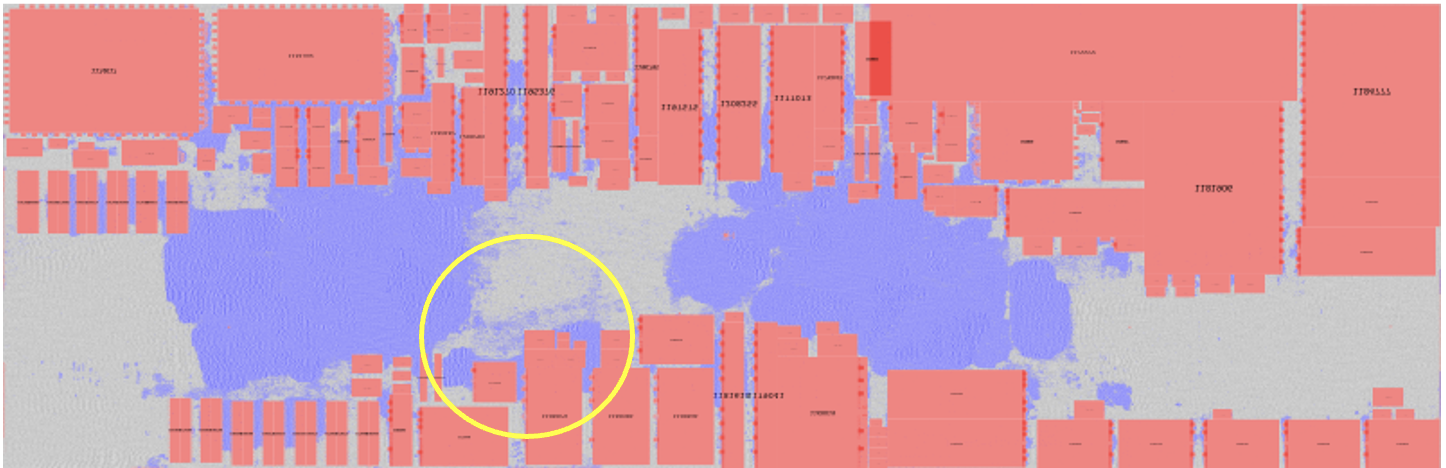}
    \subcaption{GP plot of DREAMPlace-Nesterov}
    \end{subfigure}
\caption{The global placement plot comparison of EvoStage and DREAMPlace-Nesterov~\cite{lin2020dreamplace} on superblue3, where movable standard cells are denoted in blue and fixed macro cells are denoted in red. The critical areas, where EvoStage produces better layouts (e.g., neater boundaries or stronger cohesion), are highlighted in yellow circles.}
\label{plot:superblue3}
\end{figure}

\begin{figure}[h]
\centering
    \begin{subfigure}[b]{0.48\textwidth}
    \centering
    \includegraphics[width=\textwidth]{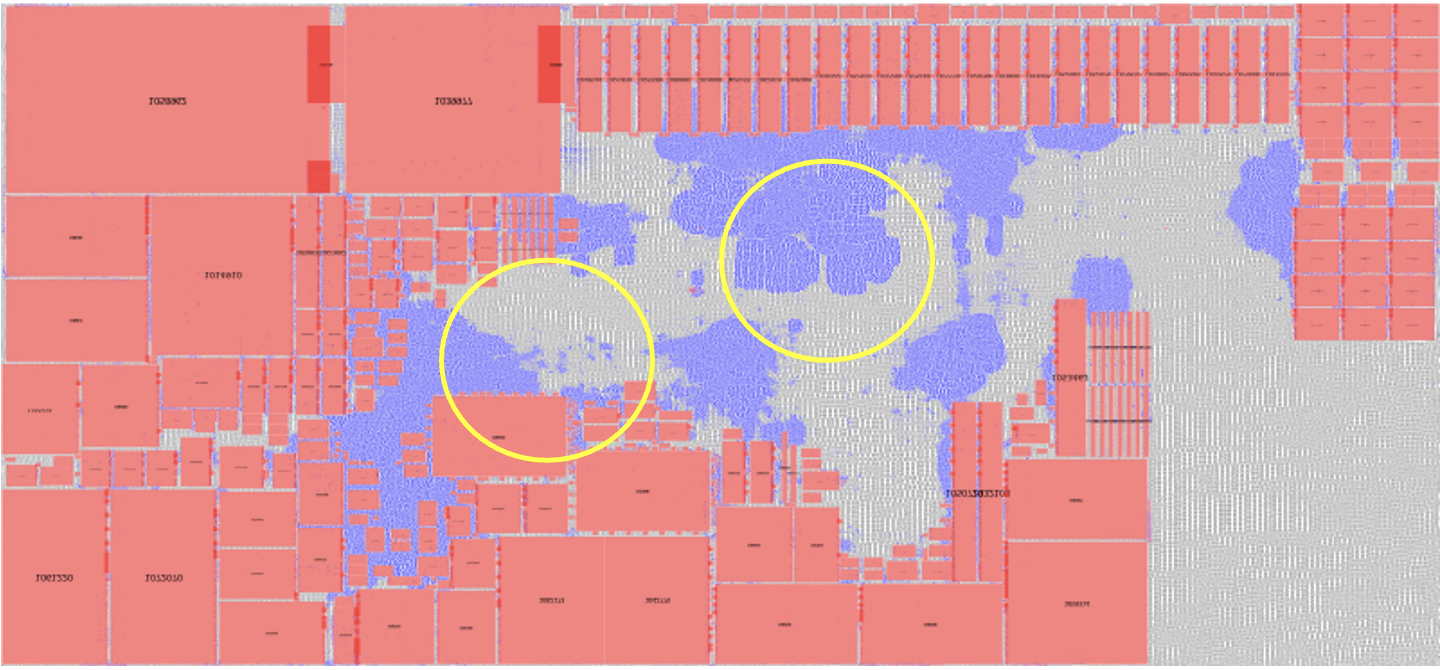} 
    \subcaption{GP plot of EvoStage}
    \end{subfigure}
    \begin{subfigure}[b]{0.48\textwidth}
    \centering
    \includegraphics[width=\textwidth]{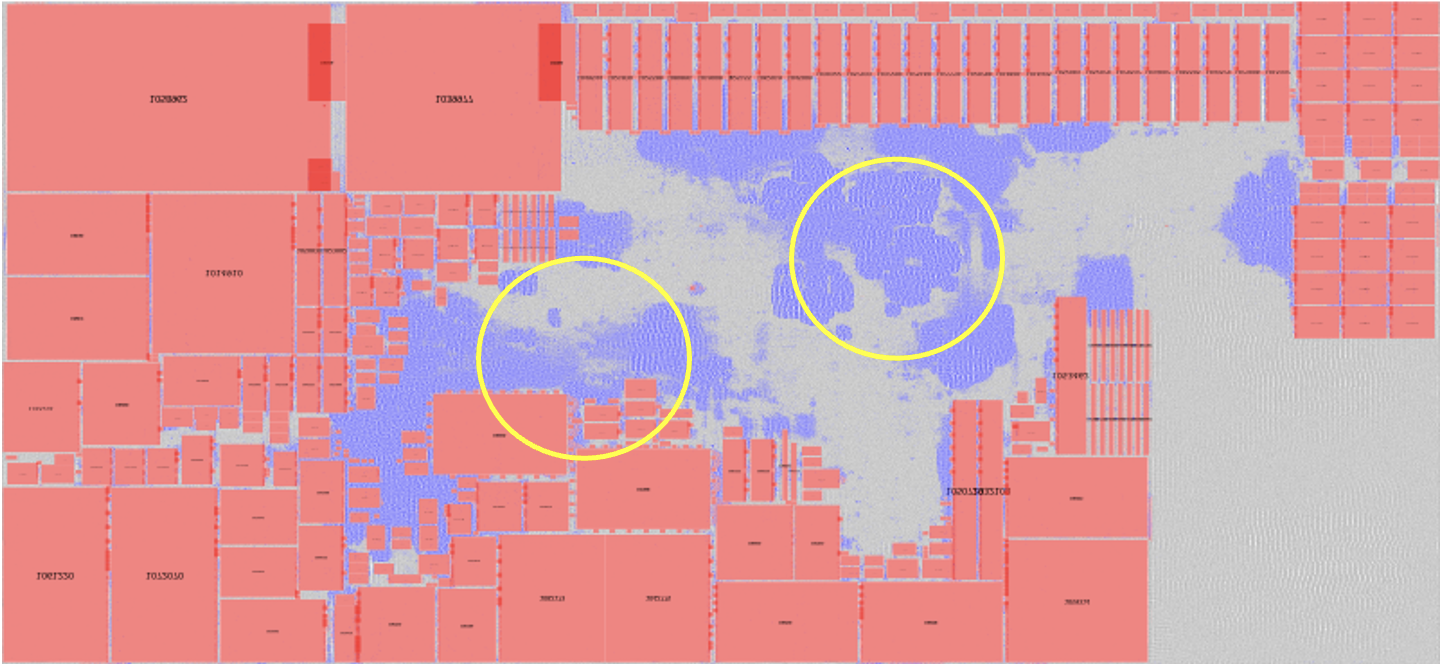}
    \subcaption{GP plot of DREAMPlace-Nesterov}
    \end{subfigure}
\caption{The global placement plot comparison of EvoStage and DREAMPlace-Nesterov~\cite{lin2020dreamplace} on superblue5, where movable standard cells are denoted in blue and fixed macro cells are denoted in red. The critical areas, where EvoStage produces better layouts (e.g., neater boundaries or stronger cohesion), are highlighted in yellow circles.}
\label{plot:superblue5}
\end{figure}

In this section, we demonstrate the intuitive comparison of the GP layout plots produced by our method EvoStage and the state-of-the-art expert-designed DREAMPlace-Nesterov~\cite{lin2020dreamplace}. We provide visualization examples of 4 chip cases where EvoStage achieves more significant improvements. As shown in Figure~\ref{plot:adaptec4} to~\ref{plot:superblue5}, EvoStage produces layouts where the standard cell clusters (shown in blue) exhibit neater boundaries and stronger cohesion, which can provide more effective guidance for upstream and downstream tasks (for instance, guiding the partitioning process in floorplanning). The critical areas, where EvoStage produces better layouts (e.g., neater boundaries or stronger cohesion), are highlighted in yellow circles. 

\end{document}